\documentclass{article}

\usepackage[nonatbib,final]{neurips_2022} 

\usepackage{multirow}
\usepackage[utf8]{inputenc} 
\usepackage[T1]{fontenc}    
\usepackage{url}            
\usepackage{booktabs}       
\usepackage{amsfonts}       
\usepackage{nicefrac}       
\usepackage{microtype}      
\usepackage{xcolor}         
\usepackage{float}
\usepackage{amsmath}
\usepackage{amsthm}
\usepackage{latexsym}
\usepackage{amssymb}
\usepackage{etoolbox}
\usepackage{array}
\usepackage{booktabs}
\usepackage{enumitem}
\usepackage{titlesec}
\usepackage{caption}
\usepackage{subcaption}
\usepackage{comment}
\usepackage{algorithm}
\usepackage{algorithmicx}
\usepackage[noend]{algpseudocode}
\usepackage{pifont}
\usepackage{verbatim,ifthen,alltt}
\usepackage{microtype}      
\usepackage{stmaryrd}
\usepackage{xfrac}
\usepackage{tikz}
\usepackage{pgf}

\usetikzlibrary{fit,calc}

\definecolor{darkslategray}{rgb}{0.18, 0.31, 0.31} 
\definecolor{platinum}{rgb}{0.9, 0.89, 0.89} 
\definecolor{gray}{rgb}{.4,.4,.4}
\definecolor{midgrey}{rgb}{0.5,0.5,0.5}
\definecolor{middarkgrey}{rgb}{0.35,0.35,0.35}
\definecolor{darkgrey}{rgb}{0.3,0.3,0.3}
\definecolor{darkred}{rgb}{0.7,0.1,0.1}
\definecolor{midblue}{rgb}{0.2,0.2,0.7}
\definecolor{darkblue}{rgb}{0.1,0.1,0.5}
\definecolor{darkgreen}{rgb}{0.1,0.5,0.1}
\definecolor{defseagreen}{cmyk}{0.69,0,0.50,0}

\def\pdfauthor{X. Huang and J. Marques-Silva}
\usepackage[pdftex%
,colorlinks=true%
,bookmarks=true%
,linkcolor=darkblue%
,citecolor=darkblue%
,urlcolor=darkblue%
,plainpages=false]{hyperref}
\hypersetup{%
pdfauthor={{\pdfauthor}},
pdftitle={{}}
}
\usepackage[noabbrev,nameinlink,capitalise]{cleveref}

\newtheorem{proposition}{Proposition}

\newtheorem{example}{Example}

\crefname{theorem}{Theorem}{Theorems}
\crefname{lemma}{Lemma}{Lemmas}
\crefname{proposition}{Proposition}{Propositions}
\crefname{definition}{Definition}{Definitions}
\crefname{corollary}{Corollary}{Corollaries}
\crefname{example}{Example}{Examples}
\crefname{claim}{Claim}{Claims}
\crefname{assumption}{Assumption}{Assumptions}
\crefname{enumi}{}{}

\newcommand{\cmark}{\mbox{\ding{51}}}

\newcommand{\spc}{$\,$}
\newcommand{\fml}[1]{{\mathcal{#1}}}

\newcommand{\tn}[1]{\textnormal{#1}}

\newcommand{\mbf}[1]{\ensuremath\mathbf{#1}}
\newcommand{\msf}[1]{\ensuremath\mathsf{#1}}
\newcommand{\mbb}[1]{\ensuremath\mathbb{#1}}
\newcommand{\mrm}[1]{\ensuremath\mathrm{#1}}

\newcommand{\waxp}{\ensuremath\mathsf{WAXp}}
\newcommand{\wcxp}{\ensuremath\mathsf{WCXp}}
\newcommand{\axp}{\ensuremath\mathsf{AXp}}
\newcommand{\cxp}{\ensuremath\mathsf{CXp}}

\DeclareMathOperator*{\sv}{\msf{Sv}}

\newcommand{\stwop}{\mrm{\Sigma}_2^\tn{P}}

\newcommand{\oper}[1]{\ensuremath\mathsf{#1}}

\newcommand{\bigland}{\ensuremath\bigwedge}

\newcommand{\relevant}{\oper{Relevant}}
\newcommand{\irrelevant}{\oper{Irrelevant}}

\newcounter{tableeqn}[table]

\DeclareMathOperator*{\lequiv}{\leftrightarrow}
\DeclareMathOperator*{\limply}{\rightarrow}

\newcommand{\jnoteF}[1]{}

\newcolumntype{L}[1]{>{\raggedright\let\newline\\\arraybackslash\hspace{0pt}}m{#1}}
\newcolumntype{C}[1]{>{\centering\let\newline\\\arraybackslash\hspace{0pt}}m{#1}}
\newcolumntype{R}[1]{>{\raggedleft\let\newline\\\arraybackslash\hspace{0pt}}m{#1}}

\tikzset{
  0 my edge/.style={densely dashed, my edge},
  my edge/.style={-{Stealth[]}},
}

\def\HiLi{\leavevmode\rlap{\hbox to \linewidth{\color{platinum}\leaders\hrule height .8\baselineskip depth .5ex\hfill}}}

\titlespacing{\section}{0pt}{*2.15}{*1.0}
\titlespacing{\subsection}{0pt}{*1.25}{*0.75}
\titlespacing{\subsubsection}{0pt}{*0.35}{*0.5}
\titlespacing{\paragraph}{0pt}{*0.1}{*0.575}

\makeatletter
\newcommand\nparagraph{%
  \@startsection{paragraph}
    {4}
    {\z@}
    {0.225ex \@plus0.225ex \@minus.125ex}
    {-1em}
    {\normalfont\normalsize\bfseries}%
}
\makeatother

\setitemize{noitemsep,topsep=0pt,parsep=0pt,partopsep=0pt}
\setenumerate{noitemsep,topsep=0pt,parsep=0pt,partopsep=0pt}
\setlist{nosep,leftmargin=0.45cm}

\providetoggle{long}
\settoggle{long}{false}

\makeatletter
\algnewcommand{\LineComment}[1]{\Statex \hskip\ALG@thistlm \(\triangleright\) #1}
\makeatother

\title{The Inadequacy of Shapley Values for Explainability}

\author{%
  Xuanxiang Huang
  \\
  University of Toulouse \\
  Toulouse, France \\
  \texttt{xuanxiang.huang@univ-toulouse.fr} \\
  \And
  Joao Marques-Silva \\
  IRIT, CNRS, France \\
  Toulouse, France\\
  \texttt{joao.marques-silva@irit.fr} \\
}

\begin{document}

\maketitle

%
\begin{abstract}
  This paper develops a rigorous argument for why the use of Shapley
  values in explainable AI (XAI) will necessarily yield provably
  misleading information about the relative importance of features for
  predictions. Concretely, this paper demonstrates that there exist
  classifiers, and associated predictions, for which the relative
  importance of features determined by the Shapley values will
  incorrectly assign more importance to features that are provably
  irrelevant for the prediction, and less importance to features that
  are provably relevant for the prediction.
  The paper also argues that, given recent complexity results, the
  existence of efficient algorithms for the computation of rigorous
  feature attribution values in the case of some restricted classes of
  classifiers should be deemed unlikely at best.
\end{abstract}
%

%
\section{Introduction} \label{sec:intro}

%
Motivated by the widespread adoption of machine learning (ML) in a
ever-increasing range of domains, explainable AI (XAI) is becoming
critical, both to build trust, but also to validate ML
models~\cite{pedreschi-acmcs19,samek-bk19,molnar-bk20}.
Some of the best-known methods of explainability can be broadly
organized into two families: those based on \emph{feature
attribution} and those based on \emph{feature selection}.

\paragraph{Feature selection.}
These methods identify sets of features (i.e.\ an explanation)
relevant for a prediction. One solution for feature selection is
model-agnostic, and is exemplified by tools such as
Anchors~\cite{guestrin-aaai18}. Another solution is model-based and
can be related with logic-based
abduction~\cite{msi-aaai22,ms-corr22}. A well-known concept in 
logic-based abduction is \emph{relevancy}~\cite{gottlob-jacm95},
i.e.\ whether a hypothesis (which represents a feature in the case of
explainability) is included in some irreducible explanation. A feature
that is not included in any irreducible explanation is deemed
\emph{irrelevant}.

\paragraph{Feature attribution \& Shapley values.}
These methods assign an \emph{importance} to each feature. Well-known 
examples include LIME~\cite{guestrin-kdd16} and
SHAP~\cite{lundberg-nips17}. For neural networks, dedicated methods
have also been
proposed~\cite{pedreschi-acmcs19,samek-bk19,molnar-bk20}. SHAP is
arguably among the most established XAI feature-attribution methods,
being based on computing an approximation of Shapley values.

Shapley values were originally proposed in the context of game
theory~\cite{shapley-ctg53}, and find widespread
use~\cite{roth-bk88,winter-hbk02}.
For more than a decade, Shapley values have been employed with the
purpose of explaining the predictions of ML
models, e.g.~\cite{kononenko-jmlr10,kononenko-kis14,zick-sp16,lundberg-nips17,jordan-iclr19,lundberg-naturemi20,taly-cdmake20,lundberg-nips20,feige-nips20,covert-aistats21,feige-iclr21,lakkaraju-nips21,covert-iclr22,giannotti-ccai22,watson-facct22,magazzeni-facct22,giannotti-eg22,giannotti-dmkd22,xie-bigdata22,giannotti-epjds22}.
In these settings, Shapley values represent often the relative
importance of features~\cite{sarkar-ijcai22}.
However, exact computation of Shapley values is 
considered unrealistic in practice, and so a number of different
approaches have been proposed for their approximate computation.
The computational complexity of computing exact Shapley values for
explainability has been studied
recently~\cite{vandenbroeck-aaai21,vandenbroeck-jair22}, confirming in
theory what in practice was already assumed.
Tractable cases have also been identified in recent
work~\cite{barcelo-aaai21,barcelo-corr21}.

\paragraph{SHAP vs.\ exact Shapley values.}
The original motivation for this work was to assess the rigor of
SHAP~\cite{lundberg-nips17} when compared with the exact computation
of Shapley values~\cite{shapley-ctg53}, as defined in SHAP's original
work~\cite{lundberg-nips17}.
Although it is in general unrealistic to compute exact Shapley values,
recent work proposed two algorithms for computing the exact Shapley
values in the case of deterministic decomposable Boolean
circuits (which include binary decision
trees)~\cite{barcelo-aaai21,barcelo-corr21}\footnote{One alternative
had been proposed in earlier work~\cite{lundberg-naturemi20}, but the
proposed algorithm has been shown to be
unsound~\cite{vandenbroeck-aaai21,vandenbroeck-jair22}.}.
(Deterministic decomposable Boolean circuits are represented as 
d-DNNF formulas~\cite{darwiche-jair02}, and so we will use the acronym
d-DNNF throughout.)
Thus, the initial goal of this work was to compare the scores obtained 
with SHAP against those obtained by exact computation of Shapley
values\footnote{Throughout the paper, the approximate results obtained
with the SHAP tool~\cite{lundberg-nips17} will be referred to as
\emph{SHAP}'s results, whereas the exact computation of Shapley values
for explainability, studied in~\cite{barcelo-aaai21,barcelo-corr21}
but proposed in earlier work~\cite{lundberg-nips17} will be referred
to as \emph{exact Shapley values}.}.

\paragraph{Explainability is \textsc{not} SHAP's game.}
Perhaps unsurprisingly, the results of SHAP essentially never matched
the Shapley values obtained with exact computation. This may somehow
be expected because the goal of SHAP is to measure the relative
importance of features and not so much to compute the exact value of 
feature importance.
As a result, we analyzed the orders of feature importance produced by
SHAP and by exact computation. Somewhat more surprisingly, we observed
that the obtained orders almost never match.
The conclusion to draw is that the feature attributions computed by
SHAP do not accurately capture neither the exact Shapley values, nor
the relative order of features imposed by the exact Shapley values.
Therefore, our experiments demonstrate that, as a tool for measuring
feature importance with a measure that relates with (exact) Shapley
values, SHAP is flawed.
The limitations of SHAP are analyzed in~\cref{ssec:shvssv}.

\paragraph{Exact Shapley values attribute misleading feature importance.}
Since SHAP ought not be used for assessing feature attribution, we
considered exploiting rigorous feature selection approaches, for
listing \emph{all} the explanations of a prediction, thereby obtaining
qualitative information about the relative importance of features for
a prediction. Concretely, we considered abductive
explanations~\cite{msi-aaai22}.
However, during this process, we observed that irrelevant features,
i.e.\ features that could be shown \emph{not} to be included in
\emph{any} explanation (among those based of feature selection),
would have non-zero Shapley values, i.e.\ those features would be
deemed of \emph{some} importance.
Even more unsettling, we observed that it could also happen that
relevant features, i.e.\ features that were used in some
explanation(s), would have a Shapley value of zero, i.e.\ those
features would be deemed of \emph{no} importance.
By putting a more concerted effort, we were able to discover that
other pitfalls of Shapley values could be identified, and that all the
pitfalls of Shapley values are not at all uncommon.
These additional pitfalls include (a) uncovering instances for which
at least one relevant feature has a Shapley value of zero and a
relevant feature has a non-zero Shapley value; and (b) uncovering
instances where there exist irrelevant features significantly better
ranked than  relevant features according to the order of features
determined by their Shapley values.

The issues above indicate that exact Shapley values for
explainability, as proposed originally in SHAP~\cite{lundberg-nips17},
and as computed exactly (for d-DNNFs) in more recent
work~\cite{barcelo-aaai21,barcelo-corr21}, do not reflect the
effective relevancy of features for explanations, and can incorrectly 
assign top-ranked importance to features that are irrelevant for a
prediction, and assign low-ranked importance to features that are
relevant for a prediction.
Another consequence of these observations is that SHAP and its
variants are not only flawed in the results they produce, but they are
also flawed in their core assumptions\footnote{%
  Naturally, a flawed approximation of a flawed concept of feature
  attribution offers no guarantees whatsoever of quality of
  approximating feature attribution.}.
The experimental confirmation of the limitations of exact Shapley
values are analyzed in~\cref{ssec:svvsaxp}.

Finally, we argue that some other definition of Shapley values, if it
were to exist and \emph{respect} feature (ir)relevancy, ought not be
computed in polynomial time in the case of d-DNNFs, unless
$\tn{P}=\tn{NP}$.

\paragraph{An alternative measure of feature importance.}
Given the inadequacy of Shapley values for explainability, the paper
proposes a simple measure of feature importance which respects feature
(ir)relevancy, i.e.\ irrelevant features are assigned a score of zero,
and relevant features are assigned a non-zero score. The measure is
based on the enumeration of all explanations based on feature
selection. This means that for complex ML models, the proposed measure
will be difficult to compute.

\paragraph{Additional contributions.}
Despite being empirically motivated, the paper outlines a principled
approach for uncovering a number of pitfalls with the use of Shapley
values in explainable AI (XAI).
To implement such a principled approach, and to obtain the results
outlined above, we devised dedicated algorithms for computing exact
Shapley values for any (discrete) function represented by a truth
table, but also for deciding feature relevancy/irrelevancy. The
proposed algorithms run in polynomial-time on the size of the truth
table.

\paragraph{Organization.}
The paper is organized as follows.
\cref{sec:prelim} introduces the definitions and notation used
throughout the paper.
\cref{sec:alt} develops alternative representations of abductive
explanations, which serve to relate exact Shapley values with these
explanations.
\cref{sec:svvsaxp} uncovers a number of links between the definition
of Shapley values and abductive explanations, and investigates a
number of possible issues that Shapley values may exhibit, which would
confirm the inadequacy of Shapley values for explainability.
\cref{sec:approach} outlines the algorithms developed for assessing
the existence of such issues.
\cref{sec:newmfi} briefly outlines an alternative measure for
assessing feature importance in classifier predictions. This section
also argues that variants of the definition of Shapley values are
unlikely to give correct results in polynomial-time, unless
$\tn{P}=\tn{NP}$.
\cref{sec:res} presents extensive evidence to the issues that Shapley
values exhibit, thereby demonstrating the inadequacy of Shapley values
for any form of explainability where any form of rigor matters.
\cref{sec:conc} summarizes the paper's contributions.

\section{Preliminaries} \label{sec:prelim}

The paper assumes basic knowledge of computational complexity, namely
the classes P and NP. (A standard reference is~\cite{arora-bk09}.)
%

\paragraph{Classification problems.}

Classification problems are defined on a set of features (or
attributes) $\fml{F}=\{1,\ldots,m\}$ and a set of classes
$\fml{K}=\{c_1,c_2,\ldots,c_K\}$.
Each feature $i\in\fml{F}$ takes values from a domain
$\mbb{D}_i$. In general, domains can be categorical or ordinal.
However, for the purposes of this paper, all features are assumed to
be Boolean, and so $\mbb{D}_i=\{0,1\}$. (However, some results are
straightforward to generalize to categorical features.)
Feature space is defined as
$\mbb{F}=\mbb{D}_1\times{\mbb{D}_2}\times\ldots\times{\mbb{D}_m}$,
which is this paper will result in $\mbb{F}=\{0,1\}^{m}$.
The notation $\mbf{x}=(x_1,\ldots,x_m)$ denotes an arbitrary point in
feature space, where each $x_i$ is a variable taking values from
$\mbb{D}_i$. The set of variables associated with features is
$X=\{x_1,\ldots,x_m\}$.
Moreover, the notation $\mbf{v}=(v_1,\ldots,v_m)$ represents a
specific point in feature space, where each $v_i$ is a constant
representing one concrete value from $\mbb{D}_i$. 
A classifier $\fml{M}$ is characterized by a (non-constant)
\emph{classification function} $\kappa$ that maps feature space
$\mbb{F}$ into the set of classes $\fml{K}$,
i.e.\ $\kappa:\mbb{F}\to\fml{K}$.
An \emph{instance} 
denotes a pair $(\mbf{v}, c)$, where $\mbf{v}\in\mbb{F}$ and
$c\in\fml{K}$, with $c=\kappa(\mbf{v})$. 
Finally, an explanation problem $\fml{E}$ is a tuple
$(\fml{M},(\mbf{v},c))$.

\paragraph{Shapley values.}

This section provides a brief overview of Shapley values. 
Shapley values were first introduced by
L.~Shapley~\cite{shapley-ctg53} in the context of game theory.
Moreover, Shapley values have been proposed for explaining the
predictions of ML models in a vast number of works, which
include~\cite{kononenko-jmlr10,kononenko-kis14,zick-sp16,lundberg-nips17,jordan-iclr19,lundberg-naturemi20,taly-cdmake20,lundberg-nips20,feige-nips20,covert-aistats21,feige-iclr21,lakkaraju-nips21,covert-iclr22,giannotti-ccai22,watson-facct22,magazzeni-facct22,giannotti-eg22,giannotti-dmkd22,xie-bigdata22,giannotti-epjds22},
among many others.
Shapley values are also discussed in a number of XAI
surveys~\cite{samek-bk19,herrera-if20,molnar-bk20,longo-if21,nguyen-air22},
in addition to a recent survey on the uses of Shapley values in
machine learning~\cite{sarkar-ijcai22}.
More importantly, in some applications the use of Shapley values can
have a direct impact on human subjects (existing references
include~\cite{jansen-dphm20,yu-tc20,withnell-bb21,inoguchi-sr21,moncada-naturesr21,baptista-aij22,alsinglawi-sr22,zhang-fo22,ladbury-go22,alabi-ijmi22,sorayaie-midm22,zarinshenas-ro22,ma-er22,wang-er22,liu-bbe22,acharya-cmpb22,lund-diagnostics22,menegaz-ieee-sp22,menegaz-ieee-jbhi23,huang-plosone23,adeoye-oo23}
among many others).
The complexity of computing Shapley values (as proposed in
SHAP~\cite{lundberg-nips17}) has been studied in recent
years~\cite{barcelo-aaai21,vandenbroeck-aaai21,barcelo-corr21,vandenbroeck-jair22}.

Throughout this section, we adapt the notation used in recent
work~\cite{barcelo-aaai21,barcelo-corr21}, which builds on the work
of~\cite{lundberg-nips17}.

Let $\upsilon:2^{\fml{F}}\to2^{\mbb{F}}$ be defined by\footnote{%
Throughout the paper, we distinguish function and predicate arguments
from their parameterizations by separating them with ';' instead of
',', as in $\upsilon(\fml{S};\mbf{v})$ or
$\phi(\fml{S};\fml{M},\mbf{v})$.},
\begin{equation}
  \upsilon(\fml{S};\mbf{v})=\{\mbf{x}\in\mbb{F}\,|\,\land_{i\in\fml{S}}x_i=v_i\}
\end{equation}
i.e.\ for a given set $\fml{S}$ of features, and parameterized by
the point $\mbf{v}$ in feature space, $\upsilon(\fml{S};\mbf{v})$
denotes all the points in feature space that have in common with
$\mbf{v}$ the values of the features specified by $\fml{S}$. 

Also, let $\phi:2^{\fml{F}}\to\mbb{R}$ be defined by,
\begin{equation}
  \phi(\fml{S};\fml{M},\mbf{v})=\frac{1}{2^{|\fml{F}\setminus\fml{S}|}}\sum\nolimits_{\mbf{x}\in\upsilon(\fml{S};\mbf{v})}\kappa(\mbf{x})
\end{equation}
Hence, given a set $\fml{S}$ of features,
$\phi(\fml{S};\fml{M},\mbf{v})$ represents the average value of the
classifier over the points of feature space represented by
$\upsilon(\fml{S};\mbf{v})$.
For the purposes of this paper, and in contrast
with~\cite{barcelo-aaai21,barcelo-corr21}, we will solely consider a
uniform distribution of the inputs, and so the dependency with the
input distribution is not accounted for.

Finally, let $\sv:\fml{F}\to\mbb{R}$ be defined by,
\begin{equation} \label{eq:defsv}
  \sv(i;\fml{M},\mbf{v})=\sum\nolimits_{\fml{S}\subseteq(\fml{F}\setminus\{i\})}\frac{|\fml{S}|!(|\fml{F}|-|\fml{S}|-1)!}{|\fml{F}|!}\left(\phi(\fml{S}\cup\{i\};\fml{M},\mbf{v})-\phi(\fml{S};\fml{M},\mbf{v})\right)
\end{equation}

Given an instance $(\mbf{v},c)$, the Shapley value assigned to each
feature measures the \emph{importance} of that feature for the given
prediction.
A positive/negative value indicates that the feature can contribute to
changing the prediction, whereas a value of 0 indicates no
contribution.
Moreover, and as our results demonstrate, SHAP never really replicates
exact Shapley values. As a result, we focus on the relative order of
features imposed by the computed Shapley values. The motivation is
that, even if the computed values are not correct (as in the case of
SHAP), then what matters for a human decision maker is the order of
features in terms of their importance for the prediction.

\paragraph{Logic-based explanations.}
Given an explanation problem $\fml{E}=(\fml{M},(\mbf{v},c))$, an
abductive explanation
(AXp)~\cite{inms-aaai19,darwiche-ecai20,msi-aaai22,ms-corr22} (which
is also referred to as a PI-explanation~\cite{darwiche-ecai20})
represents an irreducible set of features which, if fixed to the
values dictated by $\mbf{v}$, are sufficient for the prediction.
Similarly, a contrastive explanation (CXp) represents an irreducible
set of features which, if allowed to take any value from their domain
(while the other features remain fixed), allows the prediction to
change.

A weak AXp (resp.~CXp) is a subset $\fml{X}$ (resp.~$\fml{Y}$) of
$\fml{F}$ such that the following predicate(s) hold(s):
\begin{align} \label{eq:waxp}
  \waxp(\fml{X};\fml{M},\mbf{v}) & {:=}
  \forall(\mbf{x}\in\mbb{F}).
  \left[
    \bigland\nolimits_{i\in{\fml{X}}}(x_i=v_i)
    \right]
  \limply(\kappa(\mbf{x})=c) \\
  \wcxp(\fml{Y};\fml{M},\mbf{v}) & {:=}
  \exists(\mbf{x}\in\mbb{F}).
  \left[
    \bigland\nolimits_{i\in{\fml{F}\setminus\fml{Y}}}(x_i=v_i)
    \right]
  \land(\kappa(\mbf{x})\not=c)
\end{align}
If $\waxp(\fml{X})$ (resp.~$\wcxp(\fml{Y})$) holds and $\fml{X}$
(resp.~$\fml{Y}$) is irreducible, then it is an AXp 
(resp.~CXp), and we say that $\axp(\fml{X})$ (resp.~$\axp(\fml{X})$)
holds. (Clearly, the predicates $\waxp$, $\wcxp$, $\axp$ and $\cxp$
map subsets of$\fml{F}$ into $\{0,1\}$.)
AXp's and CXp's exhibit a minimal hitting set (MHS) duality, in the
sense that every AXp is a minimal 
hitting set of all the CXp's and
vice-versa~\cite{inams-aiia20,msi-aaai22,ms-corr22}.

Moreover, for an explanation problem $\fml{E}=(\fml{M},(\mbf{v},c))$,
the sets of abductive and contrastive explanations are defined as follows:
\begin{align}
  \mbb{A}(\fml{E}) & = \{ \fml{X}\subseteq\fml{F}\,|\,\axp(\fml{X};\fml{M},\mbf{v}) \} \\
  \mbb{C}(\fml{E}) & = \{ \fml{Y}\subseteq\fml{F}\,|\,\cxp(\fml{Y};\fml{M},\mbf{v}) \}
\end{align}

It should be underscored that an abductive explanation for an
explanation problem $\fml{E}$ can be viewed as a \emph{rule}:
\[
\tn{IF} \quad \land_{i\in\fml{X}}(x_i=v_i) \quad \tn{THEN} \quad
\kappa(\mbf{x})=c
\]
where we further require, invoking Occam's razor, that $\fml{X}$ be
irreducible.

Throughout the paper, we will also use extensively the concept of
feature relevancy~\cite{hcmpms-corr22,hims-aaai23,hcmpms-tacas23}.
Given an explanation problem $\fml{E}=(\fml{M},(\mbf{v},c))$, a
feature $i\in\fml{F}$ is \emph{relevant} if it is included in at least
one AXp. Formally, we say that $i$ is relevant if
$i\in\cup_{\fml{X}\in\mbb{A}(\fml{E})}\fml{X}$. If a feature is not
relevant, then it is \emph{irrelevant}.
The minimal hitting set duality of abductive and contrastive
explanations also implies that
$i\in\cup_{\fml{X}\in\mbb{A}(\fml{E})}\fml{X}$ iff
$i\in\cup_{\fml{Y}\in\mbb{C}(\fml{E})}\fml{Y}$.
Furthermore, given $\fml{E}$, the predicates
$\relevant:\fml{F}\to\{0,1\}$ and $\irrelevant:\fml{F}\to\{0,1\}$ are
defined as follows.
Given $i\in\fml{F}$, $\relevant(i)$ holds true iff $i$ is relevant,
and $\irrelevant(i)$ is given by
$\irrelevant(i)\lequiv\neg\relevant(i)$.
It is plain that, if we write all the AXp's are rules, none of the
rules will include irrelevant features.

AXp's and CXp's have been studied in recent
years~\cite{inams-aiia20,msgcin-nips20,icshms-cp20,barcelo-nips20,kutyniok-jair21,msgcin-icml21,ims-ijcai21,kwiatkowska-ijcai21,ims-sat21,mazure-cikm21,marquis-kr21,msi-aaai22,rubin-aaai22,barcelo-nips22}, 
covering their computational complexity, their application to
different families of classifiers, but also addressing a number of
generalizations. Overviews of logic-based explainability are
available~\cite{ignatiev-ijcai20,msi-aaai22,ms-corr22}.

Abductive explanations as defined above are underpinned in logic-based
abduction~\cite{gottlob-jacm95}. Moreover, abduction is often viewed
as inference to the best explanation in
philosophy~\cite{douven-sep21}.

\paragraph{Related work.}

Different works have studied apparent limitations of model-agnostic
explanations (and so of Shapley values).
Some works reported on the unsoundness of model-agnostic
explanations~\cite{inms-aaai19,ignatiev-ijcai20}. One possible
criticism of these earlier results is that some model agnostic
explainers do not aim to select subsets of features relevant to the
prediction, but target instead finding an absolute/relative order of
feature importance. 
In contrast, a number of authors have reported pitfalls with the use
of SHAP and Shapley values as a measure of feature
importance~\cite{shrapnel-corr19,friedler-icml20,najmi-icml20,taly-cdmake20,nguyen-ieee-access21,procaccia-aaai21,sharma-aies21,guigue-icml21,taly-uai21,friedler-nips21,roder-mlwa22}.
However, these earlier works do not identify fundamental flaws with
the use of Shapley values in explainability.
Attempts at addressing those pitfalls include proposals to integrate
Shapley values with abductive explanations, as reported in
recent work~\cite{labreuche-sum22}. 

\section{Alternative Representation of Abductive Explanations}
\label{sec:alt}

This section proposes a different representation of abductive
explanations, which will serve to highlight the similarities and
differences with respect to exact Shapley values.

\subsection{Relationship with Prime Implicants of Boolean Functions}

We relate AXp's with Boolean functions as follows.
For each $i\in\fml{F}$, let the Boolean variable $s_i$ be associated
with selecting feature $i$ (for inclusion in some set).
Moreover, let $\pi:\{0,1\}^m\to2^{\fml{F}}$ be defined as follows,
with $\mbf{s}=(s_1,\ldots,s_m)\in\{0,1\}^m$:
\[
\pi(\mbf{s}) = \{i\in\fml{F}\,|\,s_i=1\}
\]
As observed below, $s_i=1$ denotes that $x_i$ is to be fixed to the
value dictated by the $i^{\tn{th}}$ coordinate of $\mbf{v}$, whereas
$s_i=0$ denotes that $x_i$ is free to take any value from its domain.
Moreover, we let $\fml{S}^{\mbf{s}}=\pi(\mbf{s})$ and
$\fml{U}^{\mbf{s}}=\fml{F}\setminus\fml{S}^{\mbf{s}}$.
Finally, we define a Boolean function $\sigma:\{0,1\}^m\to\{0,1\}$,
as follows.
\begin{equation} \label{eq:defsigma1}
  \begin{array}{ccccc}
    \sigma(\mbf{s})=1 & & \equiv & &
    \forall(\mbf{x}\in\mbb{F}).
    \bigwedge_{i\in\fml{S}^{\mbf{s}}}(x_i=v_i)
    \limply (\kappa(\mbf{x})=\kappa(\mbf{v}))
    \\
  \end{array}
\end{equation}
which can be stated as:
\begin{equation} \label{eq:defsigma2}
  \begin{array}{ccccc}
    \sigma(\mbf{s})=1 & & \equiv & &
    \bigwedge_{\mbf{x}\in\upsilon(\mbf{\fml{S}^{\mbf{s}}};\mbf{v})}
    (\kappa(\mbf{x})=\kappa(\mbf{v}))
  \end{array}
\end{equation}
We can view $\sigma$ as a \emph{sufficiency function}, i.e.\
$\sigma(\mbf{s})=1$ iff the variables $s_i$ assigned value 1 in
$\mbf{s}$, and which represent fixed features in $\fml{F}$, are
sufficient for predicting $c$.
Observe that $\sigma$ allows us to look at all the possible subsets 
($\fml{S}$) of fixed features (each uniquely represented by a
different $\mbf{s}$) and, if $\sigma(\mbf{s})=1$, then we know that
the subset of fixed features encoded by $\mbf{s}$ suffices for the
prediction.

\begin{example} \label{ex:runex}
  We consider an example classifier $\fml{M}$, with
  $\fml{F}=\{1,2,3\}$, $\mbb{F}=\{0,1\}^3$, with the classification
  function
  $\kappa(x_1,x_2,x_3)=(x_1\land{x_2})\lor(\neg{x_1}\land{x_3})$.
  Furthermore, the example instances are $((1,1,1),1)$ and
  $((1,0,1),0)$.

  The computation of the $\sigma$ values and the sets of AXp's and
  CXp's are shown in~\cref{tab:sigma}. All these results are easy to
  confirm.
  As can be observed, all features are relevant for the instance
  $((1,1,1),1)$, but feature $3$ is irrelevant for the instance
  $((1,0,1),0)$ since feature 3 is not included in any AXp (or CXp).

  For the first instance, i.e.\ $((1,1,1),1)$, it is plain to conclude
  that for any feature $i$ there exists at least one AXp (or CXp) that
  contains $i$. Hence, for any feature $i$, one can provide a human
  decision maker with an irreducible set of features, that is
  sufficient for guaranteeing the prediction, which contains $i$.
  Observe that, for this instance, $\{1,2\}$ and $\{2,3\}$ are the
  AXp's, which can be interpreted as the following rules:
  \[
  \tn{IF} \quad (x_1=1)\land(x_2=1) \quad \tn{THEN} \quad
  \kappa(\mbf{x})=1
  \]
  and,
  \[
  \tn{IF} \quad (x_2=1)\land(x_3=1) \quad \tn{THEN} \quad
  \kappa(\mbf{x})=1
  \]
  
  For the second instance, i.e.\ $((1,0,1),0)$, one can observe that
  for any pick $\fml{P}$ of the features to fix, such that $\fml{P}$
  is sufficient for the prediction to be 0, there is always an
  irreducible set of features $\fml{Q}\subseteq\fml{P}$ that does
  \emph{not} contain feature 3.
  Similarly, for any pick of features that is sufficient to allow
  changing the prediction, there is always an irreducible subset that
  does not contain feature 3. Invoking Occam's razor, there is no
  reason whatsoever to propose as an explanation for the instance
  $((1,0,1),0)$ a set of features that includes feature 3.
  
  For this second instance, $\{1,2\}$ is the only AXp, which can be
  interpreted as the following rule:
  \[
  \tn{IF} \quad (x_1=1)\land(x_2=0) \quad \tn{THEN} \quad
  \kappa(\mbf{x})=0
  \]
  As can be observed, $x_3$ plays no role in the prediction.
\end{example}

\begin{table}[t]
  \begin{subtable}[b]{0.185\textwidth}
    \centering{
      \scalebox{0.925}{\renewcommand{\arraystretch}{1.25}
\renewcommand{\tabcolsep}{0.275em}
\begin{tabular}{cccc}
  \toprule[1.2pt]
  $x_1$ & $x_2$ & $x_3$ & $\kappa(\cdot)$ \\
  \midrule[0.8pt]
  0     & 0     & 0     & 0 \\
  0     & 0     & 1     & 1 \\
  0     & 1     & 0     & 0 \\
  0     & 1     & 1     & 1 \\
  1     & 0     & 0     & 0 \\
  1     & 0     & 1     & 0 \\
  1     & 1     & 0     & 1 \\
  1     & 1     & 1     & 1 \\
  \bottomrule[1.2pt]
\end{tabular}
}
    }
    \caption{Truth table} 
  \end{subtable}
  \begin{subtable}[b]{0.815\textwidth}
    \centering{
      \scalebox{0.925}{\renewcommand{\arraystretch}{1.25}
\renewcommand{\tabcolsep}{0.285em}
\begin{tabular}{ccccccccccc}
  \toprule[1.2pt]
  \multicolumn{5}{c}{~} & 
  \multicolumn{3}{c}{$(\mbf{v},c)=((1,1,1),1)$} & 
  \multicolumn{3}{c}{$(\mbf{v},c)=((1,0,1),0)$} \\ \cline{6-11}
  \multirow{1}{*}{$s_1$} &
  \multirow{1}{*}{$s_2$} &
  \multirow{1}{*}{$s_3$} &
  \multirow{1}{*}{$\fml{S}^{\mbf{s}}$} &
  \multirow{1}{*}{$\fml{U}^{\mbf{s}}$} &
  $\sigma(\mbf{s})$ & $\axp(\fml{S}^{\mbf{s}})$ & $\cxp(\fml{U}^{\mbf{s}})$
  &
  $\sigma(\mbf{s})$ & $\axp(\fml{S}^{\mbf{s}})$ & $\cxp(\fml{U}^{\mbf{s}})$
  \\
  \midrule[0.8pt]
  0 & 0 & 0 & $\emptyset$ & $\{1,2,3\}$ & 0 & 0 & 0 & 0 & 0 & 0 \\
  0 & 0 & 1 & $\{3\}$ & $\{1,2\}$       & 0 & 0 & 0 & 0 & 0 & 0 \\
  0 & 1 & 0 & $\{2\}$ & $\{1,3\}$       & 0 & 0 & 1 & 0 & 0 & 0 \\
  0 & 1 & 1 & $\{2,3\}$ & $\{1\}$       & 1 & 1 & 0 & 0 & 0 & 1 \\
  1 & 0 & 0 & $\{1\}$ & $\{2,3\}$       & 0 & 0 & 0 & 0 & 0 & 0 \\
  1 & 0 & 1 & $\{1,3\}$ & $\{2\}$       & 0 & 0 & 1 & 0 & 0 & 1 \\
  1 & 1 & 0 & $\{1,2\}$ & $\{3\}$       & 1 & 1 & 0 & 1 & 1 & 0 \\
  1 & 1 & 1 & $\{1,2,3\}$ & $\emptyset$ & 1 & 0 & 0 & 1 & 0 & 0 \\
  \bottomrule[1.2pt]
\end{tabular}
}
    }
    \caption{$\sigma$ and AXp/CXp for $(\mbf{v},c)=((1,1,1),1)$ and
      $(\mbf{v},c)=((1,0,1),0)$}
  \end{subtable}
  \caption{Explanations for
    $\kappa(x_1,x_2,x_3)=(x_1\land{x_2})\lor(\neg{x_1}\land{x_3})$}
  \label{tab:sigma}
\end{table}

There is an important relationship between $\sigma$ and WAXp's, which
we state as follows:
\begin{proposition}
  Given an explanation problem $(\fml{M},(\mbf{v},c))$, and the
  sufficiency function $\sigma$, then
  \[
  \begin{array}{ccccc}
    \sigma(\mbf{s})=1 & & \equiv & & \waxp(\fml{S}^{\mbf{s}};\fml{M},\mbf{v})
  \end{array}
  \]
\end{proposition}

The following observations are immediate: $\sigma(1,\ldots,1)=1$ and
$\sigma(0,\ldots,0)=0$ (provided $\kappa$ is not constant).
Moreover, it is plain to conclude the following,

\begin{proposition}
  $\sigma$ is monotone and up-closed.
\end{proposition}

From earlier work, we have additional important results:
\begin{proposition}
  Each prime implicant of $\sigma$ is essential, i.e.\ it must be
  included in any DNF representation of $\sigma$.
\end{proposition}

It is well-known that a Boolean function can be uniquely represented 
by the disjunction of its prime
implicants~\cite{blake-phd37,quine-amm52,quine-amm55,quine-amm59,mccluskey-bstj56,brown-bk90,crama-bk11}. A
standard name is Blake's Canonical Form (BCF) of a Boolean function.
Although the number of prime implicants can be unwieldy, we may still
be interested in assessing properties of such a function
representation.

For example, since $\sigma$ can be viewed as a sufficiency function,
its prime implicants capture the sets of features that are minimally
sufficient for the prediction. Thus, if a feature $j$ does not occur
in any prime implicant of $\sigma$ it will not be included in the
function's BCF. This is a clear indication of the irrelevancy of the 
feature for the prediction.

\subsection{Feature Relevancy \& Essential Variables}
The concept of \emph{inessential} variables in Boolean functions has
been studied in the
past~\cite{hight-tcomp71,brown-bk90,hammer-dm00,crama-bk11}. 
Building on earlier work~\cite{brown-bk90,hammer-dm00,crama-bk11}, we
say that a variable $s_i$ is inessential if,
\begin{align}
  \forall(s_1,\ldots,s_{i-1},s_{i+1},&\ldots,s_m\in\{0,1\}).\nonumber\\
  &\sigma(s_1,\ldots,s_{i-1},0,s_{i+1},\ldots,s_m)=\sigma(s_1,\ldots,s_{i-1},1,s_{i+1},\ldots,s_m)
\end{align}
Inessential variables are also referred to as
\emph{irrelevant}~\cite{hammer-dm00},
\emph{redundant}~\cite{brown-bk90},
\emph{vacuous}~\cite{brown-bk90},
or \emph{dummy}~\cite{hammer-dm00}.
Because of the connection with formal explainability, we will refer to
essential (resp.\ inessential) variables as relevant
(resp.\ irrelevant).

Moreover, we underline that the definition of \emph{irrelevant}
variable is particularly demanding, in that a variable $i$ is
irrelevant only if for any point in its domain, the value of the
function does not depend on the value of $i$.
In the case of the sufficiency function $\sigma$, this requirement
signifies that a feature is irrelevant only if for any possible pick
of the fixed features, the prediction of $c$ never depends on feature
$i$. (We will see later how relevancy of features can be computed in
practice, and investigate its relationship with Shapley values.)

Furthermore,~\cite{crama-bk11} proves that:
\begin{proposition}[Theorem~1.17 in~\cite{crama-bk11}] \label{prop:t117}
  For a Boolean function $\sigma$, defined on variables
  $S=\{s_1,\ldots,s_m\}$, the following statements are equivalent:
  \begin{enumerate}
  \item The variable $s_i$ is inessential for $\sigma$.
  \item The variable $s_i$ does not appear in any prime implicant of $\sigma$.
  \item $\sigma$ has a DNF representation in which the variable $s_i$
    does not appear.
  \end{enumerate}
\end{proposition}

The previous result is a consequence of the uniqueness (and
the properties) of the representation of $\sigma$ by its prime
implicants~\cite{blake-phd37,quine-amm52,quine-amm55,quine-amm59,mccluskey-bstj56,brown-bk90}.

In terms of the sufficiency function $\sigma$, an irrelevant variable
$s_i$ is such that one can exhibit a DNF representation of $\sigma$
that does not include $s_i$. Hence, we can represent all the sets of
fixed features that are sufficient for the prediction with a DNF that
does not include $s_i$.

The relationship between essential variables and relevancy in
abductive reasoning~\cite{selman-aaai90,gottlob-ese90,gottlob-jacm95}
is apparent, and we have the following result:

\begin{proposition}
  A feature $i$ is relevant for the instance $(\mbf{v},c)$ iff $s_i$
  is essential for $\sigma$.
\end{proposition}

Furthermore, we can relate recent work on explainability queries,
especially related with
relevancy~\cite{marquis-kr20,hiims-kr21,marquis-kr21,hcmpms-corr22},
with the problem of deciding whether a variable is essential for a
Boolean function.

Given the results above, what we have established is that we can
identify the irrelevant variables of the $\sigma$ function by deciding
the relevant features for abductive explanations. While the feature
relevancy is in general $\stwop$-complete~\cite{hiims-kr21}, there
exist tractable cases, namely decision trees~\cite{hiims-kr21}. We
will exploit those tractable cases to efficiently decide feature
relevancy.
Furthermore, this section also serves to underline that the concepts
of relevancy and of inessential variables are pervasive in the study
of boolean functions and logic-based abduction, and that the two
concepts can be related.

\section{Relating Shapley Values with Abductive Explanations}
\label{sec:svvsaxp}

This section highlights some connections between Shapley values and
abductive explanations.
This will then allow us to uncover possible aspects of explainability
that Shapley fail to capture.

\subsection{Similarities and Differences}

Using the function $\upsilon$ introduced for computing Shapley values,
one can can define a explainability function
$\xi:2^{\fml{F}}\to\{0,1\}$ as follows,  
\[
\xi(\fml{S};\fml{M},\mbf{v})=\bigwedge_{\mbf{x}\in\upsilon(\fml{S};\mbf{v})}\left(\kappa(\mbf{x})=c\right)
\]
A consequence of this definition is the following one.
\begin{proposition}
  Given an explanation problem $(\fml{M},(\mbf{v},c))$, then for any
  $\fml{S}\subseteq\fml{F}$, it is the case that
  \begin{align}
    \waxp(\fml{S};\fml{M},\mbf{v})
    & ~\lequiv~
    \xi(\fml{S};\fml{M},\mbf{v}) \label{eq:waxp2xi} \\
    \waxp(\fml{S};\fml{M},\mbf{v})
    & ~\limply~
    \left(\phi(\fml{S};\fml{M},\mbf{v})=c\right) \label{eq:waxp2phi} \\
    \waxp(\fml{S};\fml{M},\mbf{v})
    & ~\limply~
    \left(\phi(\fml{S}\cup\{i\};\fml{M},\mbf{v})-
    \phi(\fml{S};\fml{M},\mbf{v})\right)=0, ~~i\in\fml{F} \label{eq:waxp2diff}
  \end{align}
\end{proposition}
(Observe that \eqref{eq:waxp2phi} and ~\eqref{eq:waxp2diff} are fairly
easy to prove in the case of Boolean classifiers with $c=1$.)

It should also be noted that weak AXp's represent some of the sets
also considered when computing Shapley values, but only those were the
prediction remains unchanged. Furthermore, (actual) AXp's are, among
those sets, the ones that are subset-minimal.

Given the comments above, we can further observe that if $\fml{S}$ is
a weak AXp, then the contribution of $\phi(\fml{S};\fml{M},\mbf{v})$
to $\sv(i;\fml{M},\mbf{v})$ will be 0. Hence, $\sv(i;\fml{M},\mbf{v})$
will be non-zero only because of sets $\fml{S}$ that are not weak
AXp's.

Thus, whereas the exact computation of Shapley values requires analyzing
all possible subsets $\fml{S}$ of fixed features, and for each such
set, it requires computing the average value of the classifier over
the non-fixed features $\fml{F}\setminus\fml{S}$, finding an AXp
requires looking at a linear number of such subsets $\fml{S}$, and for
each such subset $\fml{S}$, the goal is to decide whether the
predicted value does not change. These observations are aligned with 
the established computational complexity of computing Shapley values
and abductive
explanations~\cite{barcelo-aaai21,vandenbroeck-aaai21,barcelo-corr21,vandenbroeck-jair22}.

\subsection{Framing the Inadequacy of SHAP’s Shapley Values}
\label{ssec:inadeq}

This section answers the question: \emph{How can one demonstrate that
exact Shapley values do not capture important explainability
information?}

To answer this question, we build on logic-based explanations, but we
exploit a fundamental property of these explanations, i.e. we will
seek to determine whether a feature is in any way usable for a point
$\mbf{v}\in\mbb{F}$ to predict $\kappa(\mbf{v})=c$.

Given an instance $(\mbf{v},c)$, we will split the features into two
sets: the features that are relevant for the prediction and the
features that are irrelevant for the prediction, according to the
definition of feature relevancy introduced earlier.

If Shapley values adequately reflect explainability information about
$(\mbf{v},c)$, it should at least be the case that (i) any irrelevant
feature $i$ is deemed to have no importance for the prediction
(i.e.\ $\sv(i)=0$); and (ii) that any relevant feature $j$ should have
some degree of importance for the prediction (i.e.\ $\sv(j)\not=0$).
These are basic issues that we would expect Shapley values to
respect. However, we will look for additional issues. Overall,
we are interested in determining whether there exist Boolean functions
for which Shapley values exhibit the following issues:

\begin{enumerate}[nosep,topsep=3.0pt,itemsep=3.0pt,label=\textbf{Q\arabic*.},ref=\small\textsf{Q\arabic*},leftmargin=0.75cm]  
\item Decide whether there can exist classifiers and instances for
  which there exists at least one feature $i$ such that, \label{en:q1}
  \[\msf{Irrelevant}(i)\land\left(\msf{Sv}(i)\not=0\right)\]
  If the answer to~\cref{en:q1} is positive, this means that Shapley
  values can assign importance to features that are irrelevant for a
  prediction. (As we clarified earlier, an irrelevant feature does not
  bear any role whatsoever, over all points in feature space, in
  changing the prediction given $\mbf{v}$.)\\
  If the answer is positive, for some function and instance, we say
  that this is an I1 issue.
\item Decide whether there exist classifiers and instances for
  which there exists at least one pair of features $i_1$ and $i_2$
  such that, \label{en:q2}
  \[
  \begin{array}{l}
    \msf{Irrelevant}(i_1)\land\msf{Relevant}(i_2)\land
    \left(|\msf{Sv}(i_1)|>|\msf{Sv}(i_2)|\right)
  \end{array}
  \]
  If the answer to~\cref{en:q2} is positive, then we have an
  unsettling situation for which irrelevant features are deemed more
  important (in terms of the overall ranking of features by their
  Shapley values) than relevant features.\\
  If the answer is positive, for some function and instance, we say
  that this is an I2 issue.
\item Decide whether there exist classifiers and instances for
  which there exists at least one feature $i$ such that, \label{en:q3}
  \[\msf{Relevant}(i)\land\left(\msf{Sv}(i)=0\right)\]
  If the answer to~\cref{en:q3} is positive, this means that Shapley
  values can assign no importance to features that are actually
  important for a prediction. (As we clarified earlier, a relevant
  feature plays some role, in at least one point in feature space in
  changing the prediction given $\mbf{v}$.)\\
  If the answer is positive, for some function and instance, we say
  that this is an I3 issue.
\item Decide whether there exist classifiers and instances for
  which there exists at least one pair of features $i_1$ and $i_2$
  such that, \label{en:q4}
  \[
  [\msf{Irrelevant}(i_1)\land\left(\msf{Sv}(i_1)\not=0\right)]
  \land
  [\msf{Relevant}(i_2)\land\left(\msf{Sv}(i_2)=0\right)]
  \]
  If the answer to~\cref{en:q4} is positive, then we are faced with the
  rather problematic situation where, in the relative order of features
  induced by their Shapley values, a relevant feature is deemed of no
  important while an irrelevant feature is deemed of some importance.
  (This would represent a serious blow to whether Shapley values can
  be trusted for assigning relative importance to features.)\\
  If the answer is positive, for some function and instance, we say
  that this is an I4 issue.
\end{enumerate}

The issues above should not be expected when analyzing Shapley values.
Indeed, it is widely accepted that Shapley values measure the
\emph{influence} of a
feature~\cite{kononenko-jmlr10,kononenko-kis14,lundberg-nips17,barcelo-aaai21,vandenbroeck-aaai21}.
Concretely,~\cite{kononenko-jmlr10} reads: ``\emph{...if a feature
has no influence on the prediction it is assigned a contribution of
0.}''
But~\cite{kononenko-jmlr10} also reads ``\emph{According to the 2nd
axiom, if two features values have an identical influence on the 
prediction they are assigned contributions of equal size. The 3rd 
axiom says that if a feature has no influence on the prediction it is
assigned a contribution of 0.}'' (In this last quote, the axioms refer
to the axiomatic characterization of Shapley values.)

\paragraph{Basic issues.}
Unfortunately, and as shown in the remainder of the paper, for any of
the issues listed above, there exist Boolean functions that exhibit
one or more of those issues.
Furthermore, even without the algorithms described in the next
section, it is fairly simple to devise Boolean functions for which
issues I1 and I3 occur.

\begin{example}
  With respect to the Boolean function of~\cref{ex:runex},
  let us consider the second instance, i.e.\ $((1,0,1),0)$, for which
  features 1 and 2 are relevant, and feature 3 is irrelevant.
  We can now use \eqref{eq:defsv} to conclude that
  $\sv(3;\fml{M},(1,0,1))=\sfrac{1}{8}$, meaning that feature 3 is
  somewhat important for the prediction. Now, this is problematic,
  since feature 3 is irrelevant for the instance $((1,0,1),0)$.
  Thus, we have uncovered one example of issue I1.

  Moreover, let us consider the first instance, i.e.\ $((1,1,1),1)$,
  for which all features are relevant.
  As before, we use \eqref{eq:defsv} to conclude that
  $\sv(1;\fml{M},(1,1,1))=0$, meaning that feature 1 is not important
  for the prediction. Again, this is fairly problematic, since feature
  1 is relevant for the instance $((1,1,1),1)$,
  Thus, we have uncovered one example of issue I3.

  The experiments in~\cref{ssec:svvsaxp} offer a more comprehensive
  picture of occurence of all the issues discussed earlier in this
  section.
\end{example}

\section{Uncovering the Inadequacy of Shapley Values for Explainability}
\label{sec:approach}

This section outlines the approach used to search for Boolean
functions that answer~\cref{en:q1,en:q2,en:q3,en:q4} positively.
The first step is to devise a procedure for the systematic enumeration
of Boolean functions and all of their possible instances.
Furthermore, we need to be able to compute (exact) Shapley values
efficiently, and also to decide feature relevancy efficiently.

\subsection{Boolean Function Search by Truth Table Enumeration}

The approach we follow is to construct a truth table for $m$ variables
(i.e.\ the features of our classifier).
Depending on the computing resources available, and in the concrete
case of Boolean functions, it is realistic to construct a truth table
for a number of variables between 30 and 40.
However, we also need to enumerate all the possible Boolean functions
of $m$ variables. The number of such functions is well-known to be
$2^{2^m}$. This imposes a practical bound on the value of $m$. In this
paper we studied Boolean functions of 3 and 4 variables, but we also
obtained results for specific Boolean functions with up to 10
variables.
Moreover, the number of instances to consider is exactly the number of
entries in the truth table. So for each Boolean function, out of a
total of $2^{2^m}$, we consider $2^m$ instances.

\subsection{Computing Shapley Values in Polynomial Time}

Given a function represented by a truth table, there exist
polynomial-time algorithms (on the size of the truth table) for
computing Shapley values of all the features.
On can proceed as follows.
When computing exact Shapley values, and for each feature, the number
of subsets to consider is $2^{m-1}$, which is polynomial on the size
of the truth table. As noted earlier, each subset
picks a set of features to fix. For each subset of fixed features, the
average value (i.e.\ the value of $\phi$) can be obtained by summing
up the values of the function over the points in feature space
consistent with the fixed features, and then dividing by the number of
such points. Since we have access to the truth table, then we can
compute each Shapley value in polynomial time (on the size of the
truth table). Moreover, and over all features, we also compute the
Shapley values in polynomial time.
The argument sketched above gives the following result.

\begin{proposition}
  For a Boolean function represented by a truth table , there exists a
  polynomial-time algorithm for computing the Shapley values (as
  defined in~\eqref{eq:defsv}).
\end{proposition}

\subsection{Deciding Feature Relevancy in Polynomial Time}

As with Shapley values, for a Boolean function $\kappa$ represented by
a truth table, there exist polynomial-time algorithms for deciding
feature relevancy.
One can proceed as follows. Given the relationship between feature
relevancy and essential variables of Boolean functions, we construct
the function $\sigma$
(see~\eqref{eq:defsigma1} and ~\eqref{eq:defsigma2}), starting from
the truth-table definition of the Boolean function $\kappa$ and the
chosen instance. As argued earlier, we need to consider $2^m$ possible
picks of features, and so we can use the same truth table as the one
used for representing the function $\kappa$, the difference being the
computed function. 
Given the $\sigma$ function, and a target feature $i$, we can decide
whether $i$ is essential as follows. We set $i$ to some value, say
1. We then enumerate all possible entries in the truth table (for
$\sigma$) with $i$ fixed. For each entry, we compare its value with
the value of the entry for which $i$ has the value 0. If the values
differ, then the feature is relevant. If the feature is not deemed
relevant after looking at the entries in the truth table for
$\sigma$, then the feature is declared irrelevant.

The argument sketched above gives the following result.

\begin{proposition}
  For a Boolean function represented by a truth table, there exists a
  polynomial-time algorithm for deciding relevancy of each feature,
  and so of all the features.
\end{proposition}

\subsection{Two Case Studies}

\cref{fig:cstudies} depicts two Boolean functions of four variables
which display issues I1, I2, I3 and I4. (These two functions were
obtained with the algorithms outlined in the previous sections.)
It is plain that \emph{any} of these functions suffices for
demonstrating the inadequacy of Shapley values for explainability.
Nevertheless, the detailed experiments presented~\cref{sec:res}
provide extensive additional evidence to such inadequacy.

\begin{figure*}[t]
  \begin{minipage}{0.35\textwidth}
    \begin{subtable}[b]{1.0\textwidth}
      \begin{center}
        \renewcommand{\tabcolsep}{0.375em}
        \renewcommand{\arraystretch}{1.075}
        \begin{tabular}{cccccc}
          \toprule
          $x_1$ & $x_2$ & $x_3$ & $x_4$ &
          $\kappa_1(\mbf{x})$ & $\kappa_2(\mbf{x})$
          \\ \midrule 
          0 & 0 & 0 & 0 & 0 & 0 \\
          0 & 0 & 0 & 1 & 0 & 0 \\
          0 & 0 & 1 & 0 & 0 & 0 \\
          0 & 0 & 1 & 1 & 0 & 0 \\
          0 & 1 & 0 & 0 & 0 & 0 \\
          0 & 1 & 0 & 1 & 0 & 0 \\
          0 & 1 & 1 & 0 & 0 & 0 \\
          0 & 1 & 1 & 1 & 0 & 0 \\
          1 & 0 & 0 & 0 & 0 & 0 \\
          1 & 0 & 0 & 1 & 0 & 0 \\
          1 & 0 & 1 & 0 & 0 & 0 \\
          1 & 0 & 1 & 1 & 1 & 1 \\
          1 & 1 & 0 & 0 & 1 & 1 \\
          1 & 1 & 0 & 1 & 0 & 0 \\
          1 & 1 & 1 & 0 & 0 & 1 \\
          1 & 1 & 1 & 1 & 0 & 1 \\
          \bottomrule
        \end{tabular}
      \end{center}
      \caption{$\kappa_1$ and $\kappa_2$ answer~\cref{en:q3} positively}
      \label{tab:cexs}
    \end{subtable}
  \end{minipage}
  \begin{minipage}{0.675\textwidth}
    \begin{subfigure}[b]{0.45\textwidth}
    \begin{center}
      \scalebox{0.75}{\begin{tikzpicture}
		\node (0) at (0, 5) [draw,fill=white,circle] {$\land$};
		\node (1) at (-1, 4) [draw,fill=white,rectangle] {$x_1$};
		\node (2) at (1, 4) [draw,fill=white,circle] {$\lor$};
		\node (3) at (1, 3) [draw,fill=white,circle] {$\land$};
		\node (4) at (2, 3) [draw,fill=white,circle] {$\land$};
		\node (5) at (0, 2) [draw,fill=white,rectangle] {$\neg x_2$};
		\node (6) at (1, 2) [draw,fill=white,circle] {$\land$};
		\node (7) at (2, 2) [draw,fill=white,rectangle] {$x_2$};
		\node (8) at (3, 2) [draw,fill=white,circle] {$\land$};
		\node (9) at (0, 1) [draw,fill=white,rectangle] {$x_3$};
		\node (10) at (1, 1) [draw,fill=white,rectangle] {$x_4$};
		\node (11) at (3, 1) [draw,fill=white,rectangle] {$\neg x_3$};
		\node (12) at (4, 1) [draw,fill=white,rectangle] {$\neg x_4$};
		\draw (0) to (1);
		\draw (0) to (2);
		\draw (2) to (3);
		\draw (2) to (4);
		\draw (3) to (5);
		\draw (3) to (6);
		\draw (4) to (7);
		\draw (4) to (8);
		\draw (6) to (9);
		\draw (6) to (10);
		\draw (8) to (11);
		\draw (8) to (12);
\end{tikzpicture}}
      \caption{$\kappa_1$ in d-DNNF}
    \end{center}
    \end{subfigure}
    \begin{subfigure}[b]{0.55\textwidth}
    \begin{center}
      \hspace*{-0.775cm}\scalebox{0.75}{\begin{tikzpicture}
		\node (0) at (0, 5) [draw,fill=white,circle] {$\land$};
		\node (1) at (-1, 4) [draw,fill=white,rectangle] {$x_1$};
		\node (2) at (1, 4) [draw,fill=white,circle] {$\lor$};
		\node (3) at (1, 3) [draw,fill=white,circle] {$\land$};
		\node (4) at (2, 3) [draw,fill=white,circle] {$\land$};
		\node (5) at (0, 2) [draw,fill=white,rectangle] {$\neg x_2$};
		\node (6) at (1, 2) [draw,fill=white,circle] {$\land$};
		\node (7) at (2, 2) [draw,fill=white,rectangle] {$x_2$};
		\node (8) at (3, 2) [draw,fill=white,circle] {$\lor$};
		\node (9) at (0, 1) [draw,fill=white,rectangle] {$x_3$};
		\node (10) at (1, 1) [draw,fill=white,rectangle] {$x_4$};
		\node (11) at (3, 1) [draw,fill=white,circle] {$\land$};
		\node (12) at (4, 1) [draw,fill=white,circle] {$\land$};
		\node (13) at (2, 0) [draw,fill=white,rectangle] {$x_3$};
		\node (14) at (3, 0) [draw,fill=white,circle] {$\lor$};
		\node (15) at (4, 0) [draw,fill=white,rectangle] {$\neg x_3$};
		\node (16) at (5, 0) [draw,fill=white,rectangle] {$\neg x_4$};
		\node (17) at (2, -1) [draw,fill=white,rectangle] {$x_4$};
		\node (18) at (3, -1) [draw,fill=white,rectangle] {$\neg x_4$};
		\draw (0) to (1);
		\draw (0) to (2);
		\draw (2) to (3);
		\draw (2) to (4);
		\draw (3) to (5);
		\draw (3) to (6);
		\draw (4) to (7);
		\draw (4) to (8);
		\draw (6) to (9);
		\draw (6) to (10);
		\draw (8) to (11);
		\draw (8) to (12);
		\draw (12) to (15);
		\draw (12) to (16);
		\draw (11) to (13);
		\draw (11) to (14);
		\draw (14) to (17);
		\draw (14) to (18);
\end{tikzpicture}}
      \caption{$\kappa_2$ in d-DNNF}
    \end{center}
    \end{subfigure}

    \medskip\smallskip

    \begin{subtable}[b]{1.0\textwidth}
      \centering
        \renewcommand{\tabcolsep}{0.5em}
        \renewcommand{\arraystretch}{1.25}
        \begin{tabular}{ccccc} \toprule
          \multirow{2}{*}{Classifier} &
          \multicolumn{4}{c}{Instance: $((1,0,0,0),0)$}\\ \cline{2-5}
          & $\irrelevant(1)$ & $\sv(1)$ &$\relevant(4)$ & $\sv(4)$ \\
          \midrule
          $\fml{M}_1$ & \cmark & 0.08(3) & \cmark & 0 \\
          $\fml{M}_2$ & \cmark & 0.125   & \cmark & 0 \\
          \bottomrule
        \end{tabular}
      \caption{Instance and features that give positive answer to~\cref{en:q3}}
    \end{subtable}
  \end{minipage}
  \caption{Case studies: classifiers $\fml{M}_1$ and $\fml{M}_2$, each
    with four features, and instance $((1,0,0,0),0)$.}
  \label{fig:cstudies}
\end{figure*}

\section{An Alternative Measure of Feature Importance}
\label{sec:newmfi}

One might argue that, by changing the definition of $\phi$, it would
be possible to devise a \emph{fixed} definition of Shapley values for 
explainability. This seems unlikely in the case of d-DNNFs. If that
were to be the case, then one would be able to compute the Shapley
values for a d-DNNF in polynomial-time, and then decide the
irrelevancy of features having a Shapley value of 0. However, deciding
feature relevancy is known to be NP-complete in the case of d-DNNF
classifiers~\cite{hcmpms-tacas23,hcmpms-corr22}. So, either the
definition of Shapley values cannot be fixed, or otherwise it is
unlikely that it can be computed in polynomial-time, unless of course
that $\tn{P}=\tn{NP}$. 

An alternative measure of feature importance is to enumerate all the
AXp's of an explanation problem, and then rank the features by their
occurrence in explanations, giving more weight to the smaller
explanations. Such a measure respects \emph{feature irrelevancy} in
that irrelevant features will have a score of 0.
A drawback of such a measure of feature importance is that it hinges
on the enumeration of all abductive explanations, and their number is
worst-case exponential. However, complete (or partial) enumeration is
possible in some cases, and in those cases, the proposed measure of
feature importance could be used.
It is left for future work the identification of other possible
drawbacks.

\section{Experiments} \label{sec:res}

The experiments are organized in two main parts. The first part,
in~\cref{ssec:shvssv}, highlights the limitations of SHAP to
approximate the order of feature importance dictated by the exact
computation of Shapley values in the case of d-DNNF's.
The second part, in~\cref{ssec:svvsaxp}, summarizes the results of
searching for functions and/or instances which
answer~\cref{en:q1,en:q2,en:q3,en:q4} positively. These results
encompass the systematic search for Boolean functions defined on $k$
variables, but also the Boolean functions evaluated
in~\cref{ssec:svvsaxp}.
It should be mentioned that more detailed experiments are included in
Appendix. Namely,~\cref{app:res01} includes additional results for the
comparison between SHAP and Shapley values, whereas~\cref{app:res02}
includes additional results for the assessment of Shapley values in
explainability. The tables presented in this section summarize those
results.

\paragraph{Experimental setup.} We consider 6 publicly available
datasets from the Penn Machine Learning
Benchmarks~\cite{Olson2017PMLB}, with Boolean features and Boolean
classes. From the datasets, d-DNNFs were generated. To ensure that
SHAP and the exact computation of Shapley values are based on the same
assumptions, we sample uniformly the d-DNNFs, to produce a (uniformly
distributed) dataset which is then used by SHAP. This step is
important because abductive explanations implicitly assume a uniform
distribution on the inputs, i.e.\ all inputs are possible and equally
likely. 

To obtain d-DNNF circuits, we first trained Read-once Decision Tree (RODT) models on the given datasets
using Orange3~\cite{JMLR:demsar13a} and then mapped the obtained RODTs into d-DNNFs.
The \emph{read-once} property is defined as: \emph{each variable is encountered at most once on each path from the root to a leaf node}.
RODTs can be encoded in linear time as d-DNNF circuits~\cite{barcelo-corr21}.
The experiments were performed on a MacBook Pro with a 6-Core Intel
Core i7 2.6 GHz processor with 16 GByte RAM, running macOS Ventura.
Finally, the algorithms outlined in~\cref{sec:approach} were
implemented in Perl\footnote{The sources are available from the
authors.}, and were similarly run on macOS Ventura.

\subsection{SHAP vs.\ Shapley Values} \label{ssec:shvssv}

For comparing SHAP~\cite{lundberg-nips17} with the exact computation
of Shapley values, we use the public distribution of SHAP\footnote{%
Available from~\url{https://github.com/slundberg/shap}.}.
As observed earlier in the paper, the experiments revealed that SHAP
never matched exactly the exact Shapley values. As a result, the
experimental evaluation focuses on computing ranking of features.

\begin{table}[t]
  \centering{
    \renewcommand{\arraystretch}{1.15}
\renewcommand{\tabcolsep}{0.725em}
\begin{tabular}{lrrrrrrrr} \toprule[1.2pt]
  \multicolumn{1}{c}{Dataset} &
  \multicolumn{1}{c}{corral} &
  \multicolumn{1}{c}{mux6} &
  \multicolumn{1}{c}{xd6} &
  \multicolumn{1}{c}{3Of9} &
  \multicolumn{1}{c}{mof3710} &
  \multicolumn{1}{c}{par5+5} &
  \multicolumn{1}{c}{Total} &
  \multicolumn{1}{c}{Fraction(\%)}
  \\
  \midrule[0.8pt]
  \#$\,\tn{Err}=0$    &  0 & 50 & 0   &  63 & 0    & 0 &
  113 & 3.53 \\
  \#$\,\tn{Err}\le2$  & 20  & 64 & 14  & 312 & 0    & 66 & 
  492 & 15.38 \\
  \#$\,\tn{Err}\le4$  & 60  & -- & 64  & 452 & 0    & 301 &
  941 & 29.41 \\
  \#$\,\tn{Err}\le8$  & 64  & -- & 297 & 510 & 200  & 855 &
  2254 & 70.44 \\
  \#$\,\tn{Err}\le12$ & --  & -- & 434 & 512 & 944  & 1016 &
  3034 & 94.81 \\
  \#$\,\tn{Err}\le16$ & --  & -- & 504 &  -- & 1024 & 1024 &
  3192 & 99.75 \\
  \#$\,\tn{Err}\le20$ & --  & -- & 510 &  -- &  --  & -- &
  3198 & 99.94 \\
  \#$\,\tn{Err}\le24$ & --  & -- & 511 &  -- &  --  & -- &
  3199 & 99.97 \\
  \#$\,\tn{Err}\le32$ & --  & -- & 512 &  -- &  --  & -- &
  3200 & 100.00 \\
  \midrule[0.8pt]
  \# instances       & 64 & 64 & 512 & 512 & 1024 & 1024 &
  3200 & \\
  \bottomrule[1.2pt]
\end{tabular}

  }
  \caption{Distribution of errors (i.e.\ wrong pairs) in the SHAP
    ranking of feature importance}
  \label{tab:shvssv}
\end{table}

\cref{tab:shvssv} summarizes the cumulative error distribution for
the six datasets considered. For each dataset, we show the number of
pairs of SHAP values assigned to features which have an incorrect
order according to their exact Shapley values; these are referred to
as the wrong pairs, and the total measures the number of comparisons
to get the order given by exact Shapley values.
As can be concluded, for the vast majority of instances, the number
of wrong pairs is not zero, and a significant number of wrong pairs
can be observed for some instances.
Overall, SHAP computed the same order of features as the one computed
with exact Shapley values for only 3.53\% of the instances. For more
than 70\% of the instances, the number of wrong pairs exceeds 4. For
close to 30\% of the instances, the number of wrong pairs exceeds 8. 

Given the results above, one should not expect that, in practice, SHAP
to approximate with rigor the order of features imposed by exact
Shapley values.
Furthermore, and as shown in the next section, even the goal of 
approximating exact Shapley values may prove inadequate for
explainability.

\subsection{Shapley Values vs.\ Feature Relevancy} \label{ssec:svvsaxp}

\begin{table*}
  \begin{subtable}[c]{0.4075\textwidth}
    \centering{
      \scalebox{0.925}{\renewcommand{\arraystretch}{1.125}
\renewcommand{\tabcolsep}{0.275em}
\begin{tabular}{lr}
  \toprule
  \# of functions & 65536 \\
  \# number of instances & 1048576 \\
  \midrule
  \# of I1 issues & 781696 \\
  \# of functions with I1 issues & 65320 \\
  \% I1 issues / function & 99.67 \\
  \midrule
  \# of I2 issues & 105184 \\
  \# of functions with I2 issues & 40448 \\
  \% I2 issues / function & 61.72 \\
  Max \# of swaps & 3 \\
  Avg \# of swaps & 1.36 \\
  \midrule
  \# of I3 issues & 43008 \\
  \# of functions with I3 issues & 7800 \\
  \% I3 issues / function & 11.90 \\
  \midrule
  \# of I4 issues & 5728 \\
  \# of functions with I4 issues & 2592 \\
  \% I4 issues / function & 3.96 \\
  \bottomrule
\end{tabular}
}
    }
    \caption{Statistics for functions with 4 variables}
    \label{stab:bf-stats}
  \end{subtable}
  \begin{minipage}{0.5925\textwidth}
    \begin{subtable}[t]{1.0\textwidth}
      \centering{
        \scalebox{0.925}{\renewcommand{\arraystretch}{1.15}
\renewcommand{\tabcolsep}{0.275em}
\begin{tabular}{lrrrrrr} \toprule
  \multicolumn{1}{c}{Dataset} &
  \multicolumn{1}{c}{corral} &
  \multicolumn{1}{c}{mux6} & 
  \multicolumn{1}{c}{3of9} &
  \multicolumn{1}{c}{xd6} &
  \multicolumn{1}{c}{mof3710} &
  \multicolumn{1}{c}{par55}\\
  \midrule
  Total features & 6 & 6 & 9 & 9 & 10 & 10 \\
  Used features & 6 & 6 & 9 & 9 & 10 & 10 \\
  \# of instances & 64 & 64 & 512 & 512 & 1024 & 1024 \\
  \midrule
  \# of I1 issues & 96 & 118 & 2336 & 2256 & 2872 & 2461 \\
  \midrule
  \# of I2 issues & 16 & 38 & 290 & 288 & 32 & 331 \\
  \% \#I2 / \#I1 & 16.67 & 32.20 & 12.41 & 12.77 & 1.11 & 13.45 \\
  \bottomrule
\end{tabular}      
}
      }
      \caption{Statistics for d-DNNFs generated from datasets}
      \label{stab:sumup0}
    \end{subtable}

    \medskip

    \begin{subtable}[b]{1.0\textwidth}
      \centering{
        \scalebox{0.925}{\renewcommand{\arraystretch}{1.15}
\renewcommand{\tabcolsep}{0.275em}
\begin{tabular}{lrrrrrr} \toprule
  \multicolumn{1}{c}{Dataset} &
  \multicolumn{1}{c}{corral} &
  \multicolumn{1}{c}{mux6} & 
  \multicolumn{1}{c}{3of9} &
  \multicolumn{1}{c}{xd6} &
  \multicolumn{1}{c}{mof3710} &
  \multicolumn{1}{c}{par55} \\
  \midrule
  \%{\spc}Out{\spc}of{\spc}Order &
  25.0 & 59.4 & 56.6 & 56.3 & 3.1 & 32.3 \\
  \%{\spc}$\exists${\spc}R{\spc}in{\spc}Bot$K$ &
  25.0 & 59.4 & 36.7 & 56.3 & 0.0 & 32.3 \\
  \%{\spc}Maj{\spc}R{\spc}in{\spc}Bot$K$ &
  25.0 & 59.4 & 15.2 & 35.7 & 0.0 & 22.8 \\
  \%{\spc}$\exists${\spc}I{\spc}in{\spc}Top$K$ &
  25.0 & 0.0 & 30.9 & 45.7 & 0.0 & 0.0 \\
  \%{\spc}Maj{\spc}I{\spc}in{\spc}Top$K$ &
  25.0 & 0.0 & 1.2 & 25.8 & 0.0 & 0.0 \\
  \bottomrule
\end{tabular}      
}
      }
      \caption{Results for d-DNNFs generated from datasets}
      \label{stab:sumup}
    \end{subtable}
  \end{minipage}
  \caption{Summary of results. For~\cref{stab:sumup},
    \%{\spc}Out{\spc}of{\spc}Order denotes the percentage of instances
    with out of order irrelevant features, i.e.\ there exist some
    irrelevant feature(s) with a Shapley value (or score) greater than
    the Shapley value of some relevant feature(s);
    \%{\spc}$\exists${\spc}R{\spc}in{\spc}Bot$K$ denotes the
    percentage of instances with relevant features having some of the
    $K$ smallest scores (and which have at least one irrelevant
    feature that has a larger score);
    \%{\spc}Maj{\spc}R{\spc}in{\spc}Bot$K$ denotes the percentage of
    instances with relevant features representing the majority of the
    $K$ smallest scores (and which have at least one irrelevant
    feature that has a larger score);
    \%{\spc}$\exists${\spc}I{\spc}in{\spc}Top$K$ denotes the
    percentage of instances with irrelevant features having some of
    the $K$ largest scores (and which have at least one relevant
    feature that has a smaller score);
    \%{\spc}Maj{\spc}I{\spc}in{\spc}Top$K$ denotes the percentage of
    instances with irrelevant features representing the majority of
    the $K$ largest scores (and which have at least one relevant
    feature that has a smaller score);
    $K$ is 2 for datasets with 6 original features, and 3 for the
    other datasets.}
  \label{tab:res}
\end{table*}

\paragraph{Boolean functions with four variables.}
\cref{stab:bf-stats} summarizes the results over all Boolean functions
of 4 variables. As can be readily observed, for almost all of Boolean
functions of 4 variables (i.e.\ 99.67\% of the functions) there exists
at least one instance such that an I1 issue is observed, i.e.\ an
irrelevant feature with non-zero Shapley value.
As discussed earlier, one could argue that an irrelevant feature with
a non-zero Shapley value would still be acceptable, as long as it is
the feature with the smallest score (in absolute value).
As can be observed, for 61.75\% of the functions, the fact that an
irrelevant feature has a non-zero Shapley value (or a relevant feature
has a Shapley value of zero) also means that an irrelevant feature is
better ranked than a relevant feature. Clearly, this can induce human
decision makers in error. 
Moreover, for 11.9\% of the functions there exists at least one
instance such that an I3 issue is observed, i.e.\ a relevant 
feature with a Shapley value of 0.
Finally, for close to 4\% of the functions there exists at least one
instance such that an I4 issue is observed, i.e.\ issues I1 and I2 are
simultaneously observed for the same instance.
Observe that the analysis only focuses on classifying features as
irrelevant or irrelevant, and still a large percentage of the
functions exhibit one or more issues. It should also be noted that for
functions with more variables, similar observations were made, but in
those cases the number of functions considered was only a fraction of
the total number of functions.

One additional test that we considered was to identify the functions
and the instances for which all the irrelevant features had larger
absolute-value scores that all the relevant features. For Boolean
functions with 4 variables, we could identify more than 1500 of these
cases (out of more than 1 million instances).
For example, for the function 0110101111111111 (where the first digit
corresponds to the prediction for the point 0000, and the last digit
corresponds to the prediction for the point 1111), and for the
instance $((0,0,0,1),1)$, feature 1 is irrelevant and exhibits a 
Shapley value $\sv(1)=-0.17188$. The other features are relevant and
exhibit Shapley values $\sv(2)=\sv(3)=\sv(4)=0.11979$.
As can be observed, the Shapley values of the relevant features are
substantially smaller (in absolute value) than the Shapley value of
the irrelevant feature.

\paragraph{Results for d-DNNFs constructed from datasets.}
%
\cref{stab:sumup0} summarizes the occurrence of issues I1 and I2
over all instances for each of the d-DNNFs obtained from the
considered datasets.
(For the generated d-DNNFs, we did not observe neither I3 nor I4
issues, but these could be observed by exhaustive enumeration of the
6-variable Boolean functions.)
For each d-DNNF, the total number of I1 issues far exceeds the number
of instances. The number of I2 issues is smaller, but not negligible.
Moreover, for most of the d-DNNFs, more than 10\% of the I1 issues
yield I2 issues. For one d-DNNF, this number exceeds 30\%.

\cref{stab:sumup} summarizes some additional statistics for the
datasets and instances considered.
As can be observed, and for a very significant number of
instances, there exist irrelevant features better ranked than relevant
features (i.e.\ an instantiation of issue I2). For example, and for
three of the datasets, more than 50\% of the instances exhibit this
issue.
Moreover, and without exceptions, for all the datasets one can either
observe relevant features ranked among the smallest scores, or
irrelevant features ranked among the largest scores, or both.
Furthermore, for some datasets, and for a non-negligible number of
instances, either the relevant features represent the majority of the
lowest ranked features, or the irrelevant features represent the
majority of the highest ranked features, or both.

Among many other similar examples, we analyze one concrete instance,
namely 
$((0,0,0,1,0,1,0,0,0), 0)$
for the \texttt{xd6} dataset.
In this case, the exact Shapley values give the following ranking
(sorted by increasing value):
$\langle7,3,2,9,8,1,4,6,5\rangle$.
However, features 6 and 4 are irrelevant.
Thus, there are six \emph{relevant} features (i.e.\
$7,3,2,9,8,1$) with the \emph{lowest} Shapley values, and two
\emph{irrelevant} features (i.e.\ $4,6$) ranked among those with the
\emph{highest} Shapley values.
Evidently, the information provided by the Shapley values in this
case would
mislead a human decision maker into looking at irrelevant features,
and possible overlook relevant features.
As confirmed by~\cref{stab:sumup} in the case of the \texttt{xd6}
dataset, for 25.8\% of the instances the majority of the top-ranked
features are irrelevant.

\paragraph{Discussion.}
The results in this section offer thorough evidence to the inadequacy
of Shapley values for distinguishing between relevant features (which 
occur in some explanation) and irrelevant features (which are excluded
from any irreducible explanation).
The experiments also demonstrated that irrelevant features can be
assigned Shapley values which incorrectly give those features crucial
importance to the prediction.
In light of the results presented in this section, we conclude that
feature attribution based on exact Shapley values (as well as any
other approach that correlates in practice with exact Shapley values)
will almost surely mislead human decision makers in some use cases.

\section{Conclusions} \label{sec:conc}

For more than a decade Shapley values have represented one of the most
visible approaches for feature attribution in explainability.
Almost without exception, and motivated by its computational
complexity~\cite{vandenbroeck-aaai21,barcelo-aaai21,barcelo-corr21,vandenbroeck-jair22},
existing work approximates the computation of exact Shapley values.
This means that the adequacy of Shapley values for explainability has
not been investigated with rigor.
This paper demonstrates that exact Shapley values can attribute
incorrect importance to features. Concretely, the paper demonstrates
that there exist functions and instances such that: (i) there exist
features that are irrelevant for the prediction, but have non-zero
Shapley values; (ii) there exist pairs of features, one relevant and
the other irrelevant for the prediction, and such that the irrelevant
feature is deemed more important given its Shapley value; (iii) there
exist features which are relevant for the prediction, but which are
deemed for having no importance according to their Shapley value,
i.e.\ a Shapley value of 0; and, finally, (iv) there exist pairs of
features such that one is relevant for the prediction, but has no
importance according to its Shapley value, and the other is irrelevant
for the prediction, but has some importance according to its Shapley
value.
The conclusion from the results in this paper is that Shapley values
are not guaranteed to bear any correlation with the actual relevancy
of features for classifiers' predictions.
The significance of our results should be framed in light of the rapid
growth of practical uses of explainability methods based on Shapley
values, with one concrete example being the medical domain, of
which~\cite{jansen-dphm20,yu-tc20,withnell-bb21,inoguchi-sr21,moncada-naturesr21,baptista-aij22,alsinglawi-sr22,zhang-fo22,ladbury-go22,alabi-ijmi22,sorayaie-midm22,zarinshenas-ro22,ma-er22,wang-er22,liu-bbe22,acharya-cmpb22,lund-diagnostics22,huang-plosone23,adeoye-oo23}
represent a fraction of the many existing examples.

Furthermore, and given the results in this paper, the use of Shapley
values as a measure of feature importance should be expected to
mislead decision makers when assessing the features that impact some
prediction.
Finally, the paper proposes an alternative measure of feature
importance, which respects feature relevancy, and which is expected to
be efficient to compute in some settings, e.g. decision trees, when
using contrastive explanations as the basis for computing feature
importance~\cite{hiims-kr21,iims-jair22}.

\paragraph{Acknowledgments.}
This work was supported by the AI Interdisciplinary Institute ANITI,
funded by the French program ``Investing for the Future -- PIA3''
under Grant agreement no.\ ANR-19-PI3A-0004,
and
by the H2020-ICT38 project COALA ``Cognitive Assisted agile
manufacturing for a Labor force supported by trustworthy Artificial
intelligence''.
This work was motivated in part by discussions with several colleagues
including L.~Bertossi, A.~Ignatiev, N.~Narodytska, M.~Cooper, Y.~Izza,
R.\ Passos, J.\ Planes and N.~Asher.
%
%
%
JMS also acknowledges
the incentive provided by the ERC who, by not funding this research
nor a handful of other grant applications between 2012 and 2022, has
had a lasting impact in framing the research presented in this paper.

%

\newtoggle{mkbbl}

\settoggle{mkbbl}{false}

\iftoggle{mkbbl}{
  \bibliographystyle{abbrv}
  \bibliography{refs,xtra}
}{

}
%
%
\appendix

\section{Appendix}

\subsection{Additional Results for~\cref{ssec:shvssv}}
\label{app:res01}

\begin{figure}[t]
  \begin{center}
    \includegraphics[width=.475\textwidth]{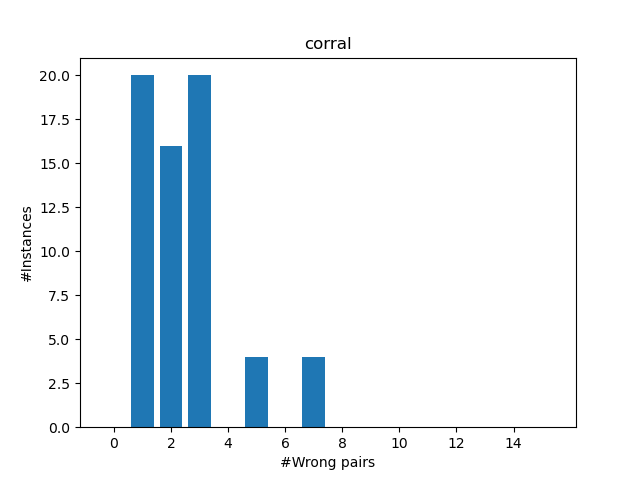}
    \quad
    \includegraphics[width=.475\textwidth]{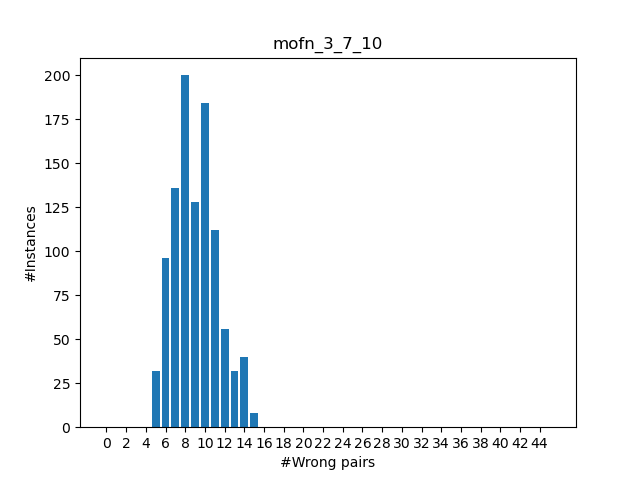} \\
    \includegraphics[width=.475\textwidth]{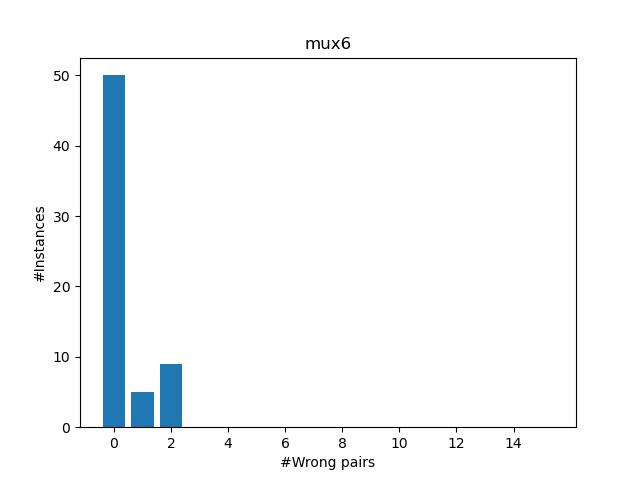}
    \quad
    \includegraphics[width=.475\textwidth]{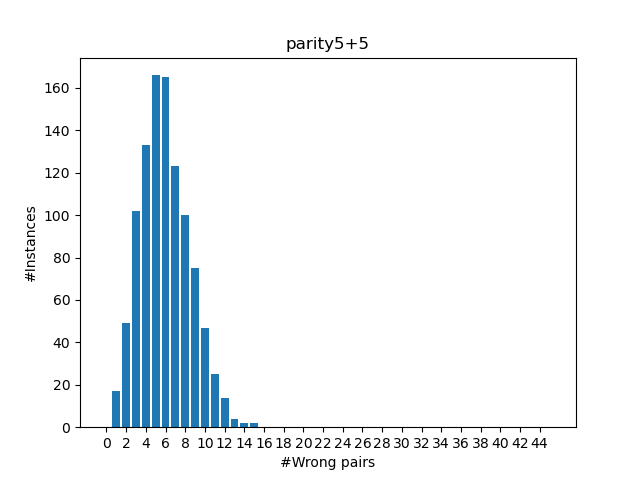} \\
    \includegraphics[width=.475\textwidth]{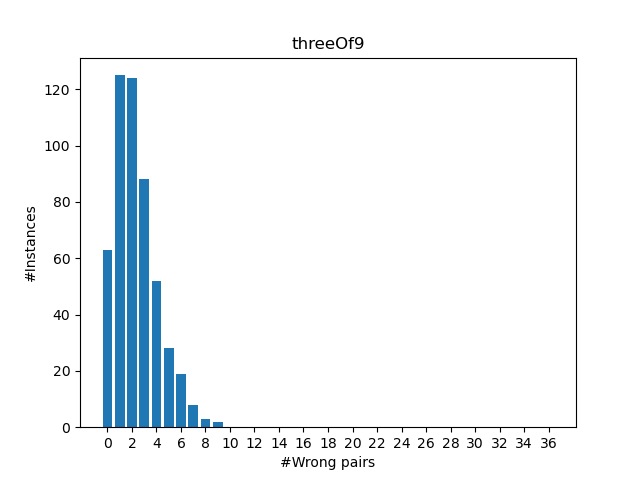}
    \quad
    \includegraphics[width=.475\textwidth]{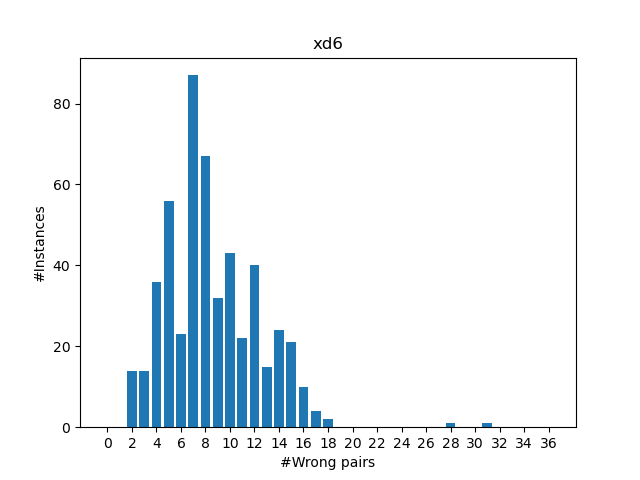}
    \caption{Distribution of wrong feature pairs with SHAP}
    \label{fig:wrong-shap}
  \end{center}
\end{figure}

\cref{fig:wrong-shap} depicts the distribution of wrong pairs between
the orders of feature importance obtained with exact Shapley values
and SHAP for the datasets considered in the experiments.

\cref{fig:wrong-shap} depicts the distribution of errors of SHAP
against exact Shapley values. As noted in~\cref{ssec:shvssv}, the
number of instances for which no error is observed is 3.53\%.

\subsection{Additional Results for~\cref{ssec:svvsaxp}}
\label{app:res02}

\begin{figure}[ht]
  \begin{center}
    \includegraphics[width=.4825\textwidth]{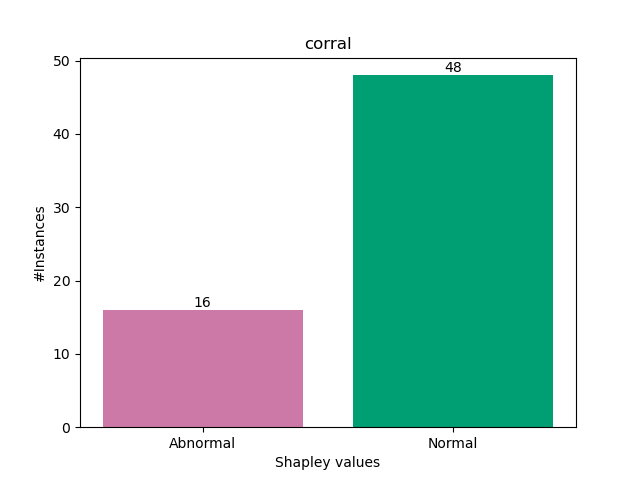}
    \quad
    \includegraphics[width=.4825\textwidth]{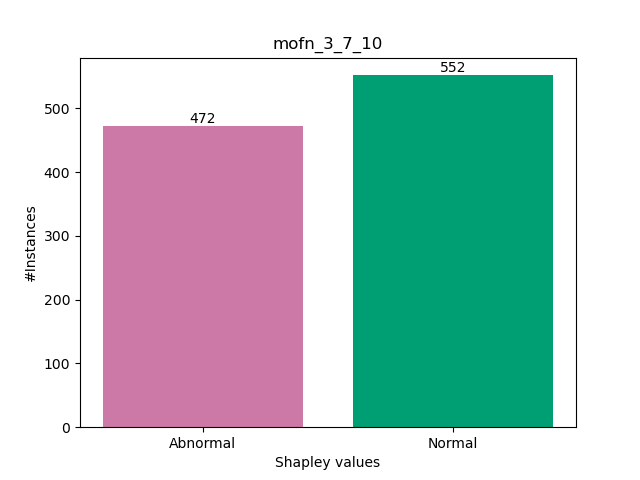} \\
    \includegraphics[width=.4825\textwidth]{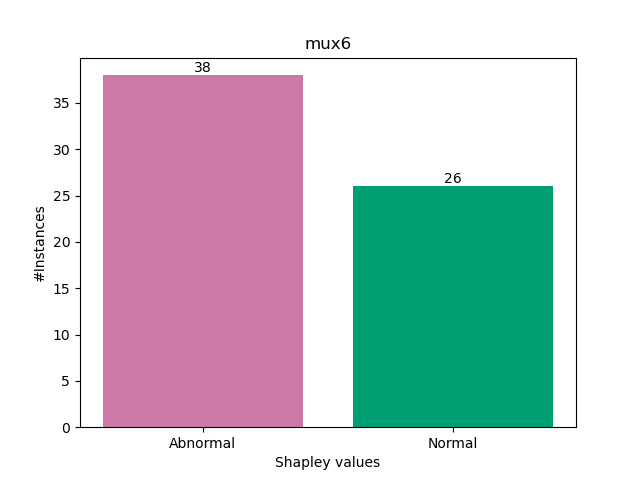}
    \quad
    \includegraphics[width=.4825\textwidth]{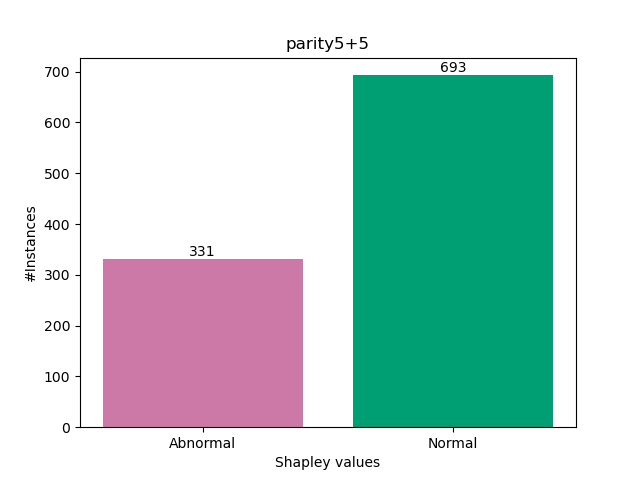} \\
    \includegraphics[width=.4825\textwidth]{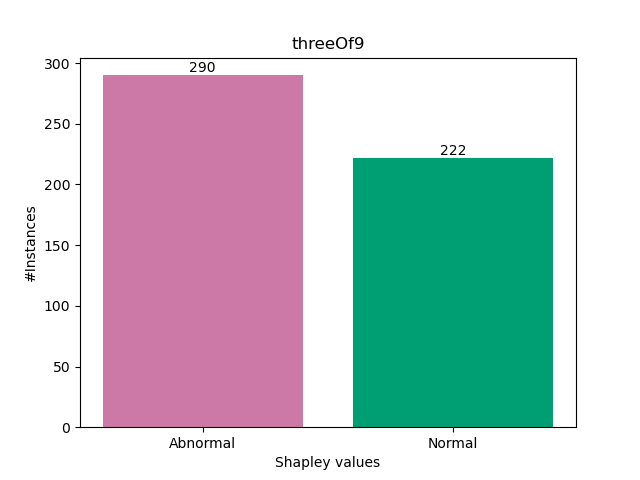}
    \quad
    \includegraphics[width=.4825\textwidth]{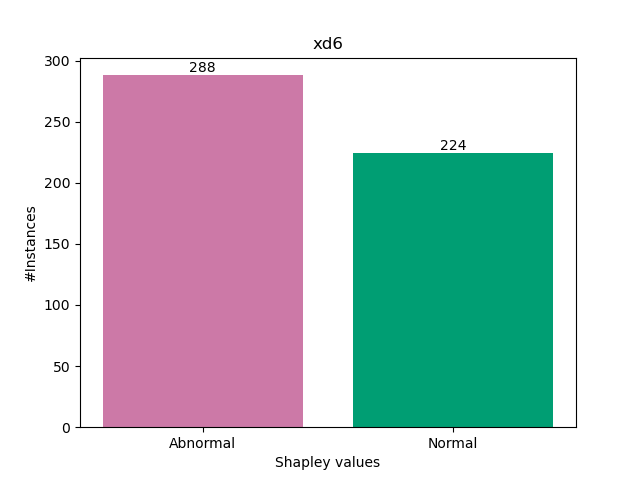}
    \caption{%
      Number of instances where the maximum Shapley value of irrelevant
      features exceeds the minimum Shapley value of
      relevant features (shown in pink), and number of instances where
      irrelevant features all have a score smaller than any relevant
      feature.}
    \label{fig:sv_irr_rel_diff}
  \end{center}
\end{figure}

\begin{figure}[ht]
  \begin{center}
    \includegraphics[width=.4825\textwidth]{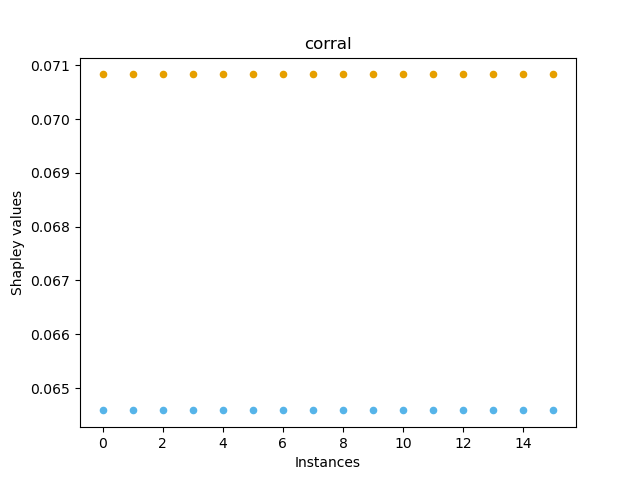}
    \quad
    \includegraphics[width=.4825\textwidth]{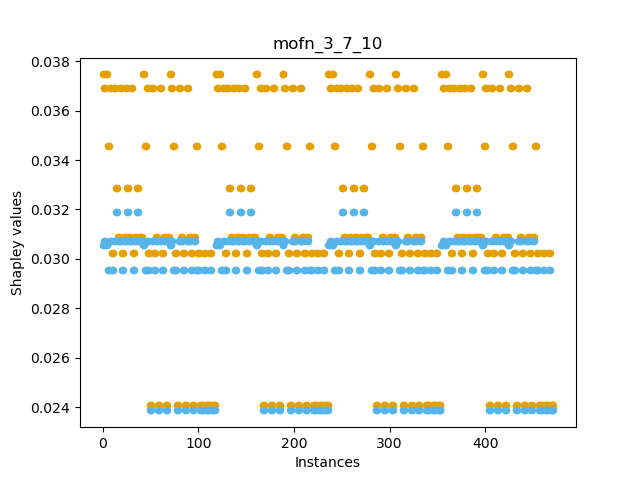} \\
    \includegraphics[width=.4825\textwidth]{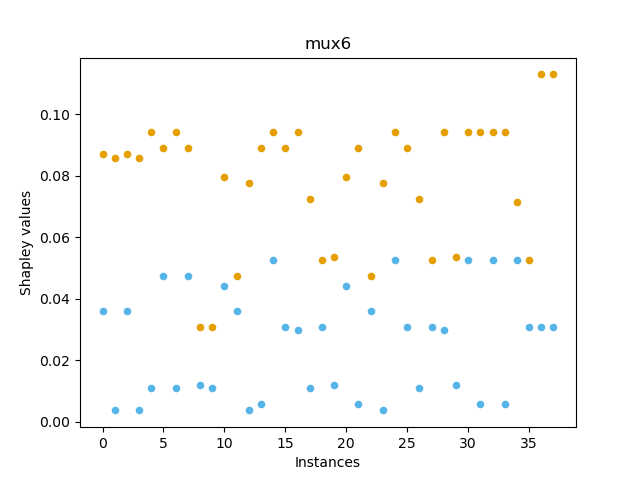}
    \quad
    \includegraphics[width=.4825\textwidth]{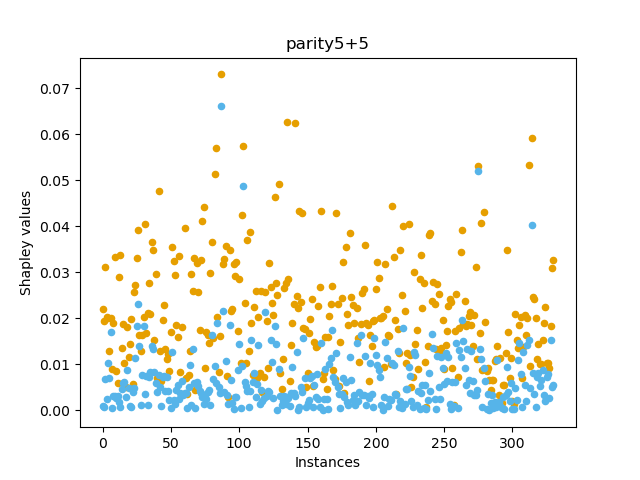} \\
    \includegraphics[width=.4825\textwidth]{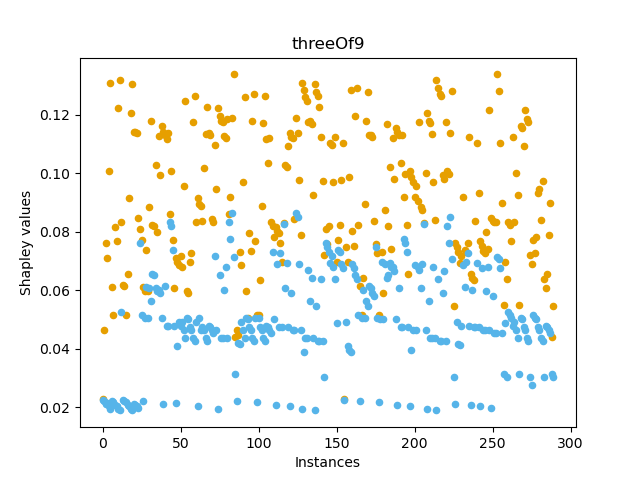}
    \quad
    \includegraphics[width=.4825\textwidth]{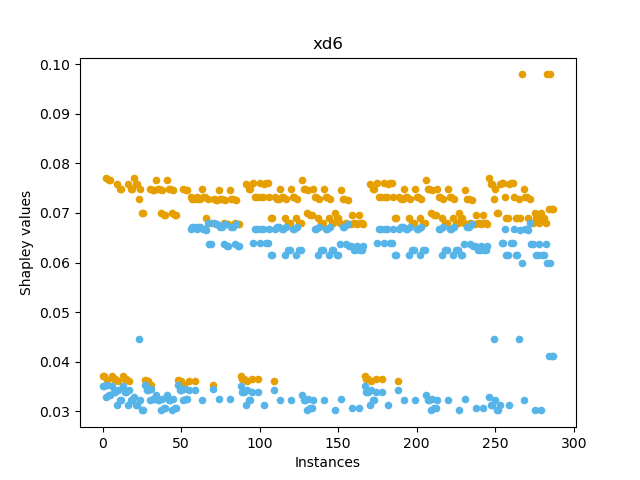}
    \caption{%
      Comparison of the maximum Shapley value of irrelevant
      features (dots in yellow) against the minimum Shapley value of
      relevant features (dots in blue). Only cases where there exist
      irrelevant features with higher scores than relevant features
      are shown.}
    \label{fig:sv_irr_rel}
  \end{center}
\end{figure}

\begin{figure}[ht]
  \begin{center}
    \includegraphics[width=.475\textwidth]{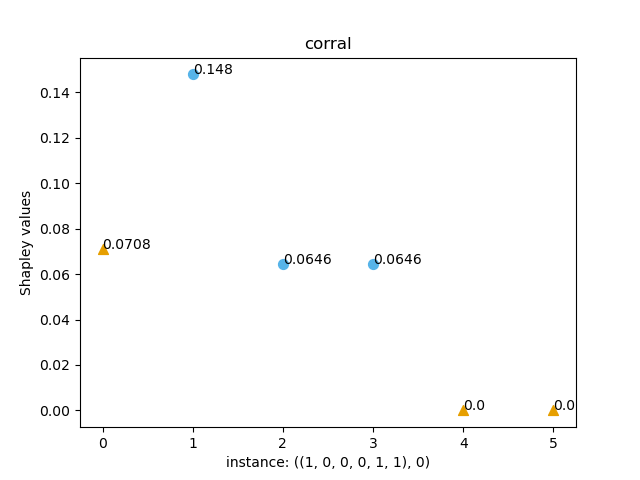}
    \quad
    \includegraphics[width=.475\textwidth]{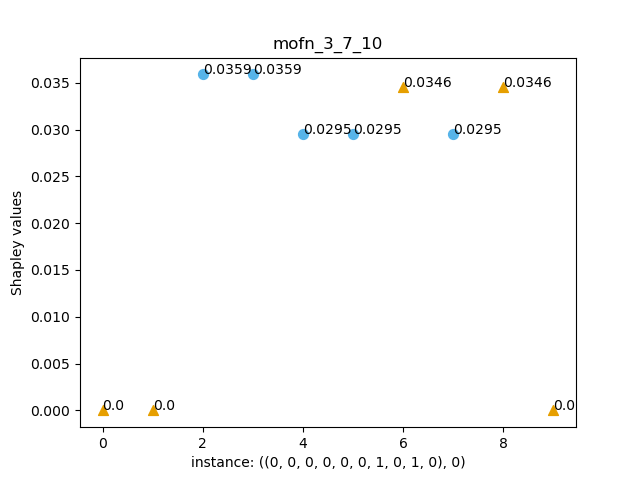} \\
    \includegraphics[width=.475\textwidth]{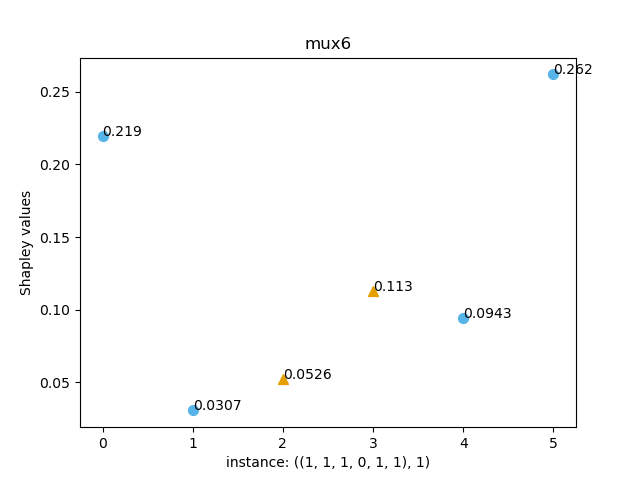}
    \quad
    \includegraphics[width=.475\textwidth]{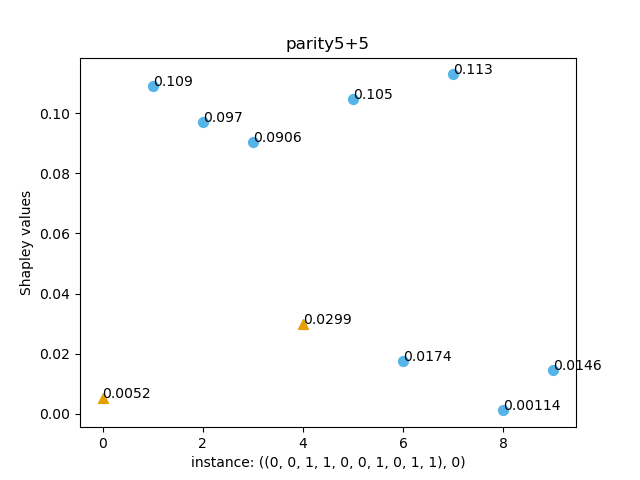} \\
    \includegraphics[width=.475\textwidth]{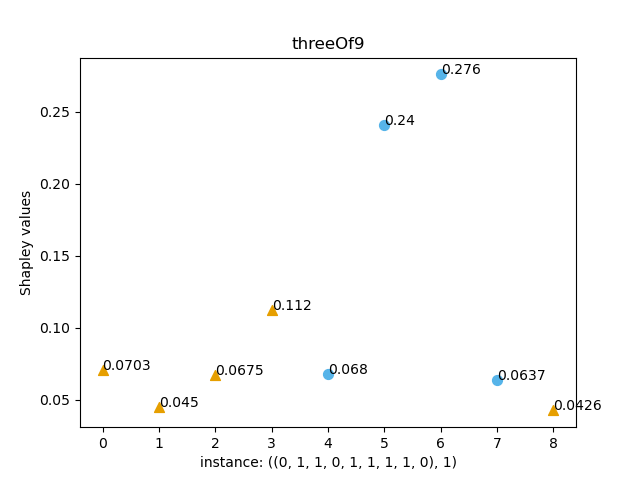}
    \quad
    \includegraphics[width=.475\textwidth]{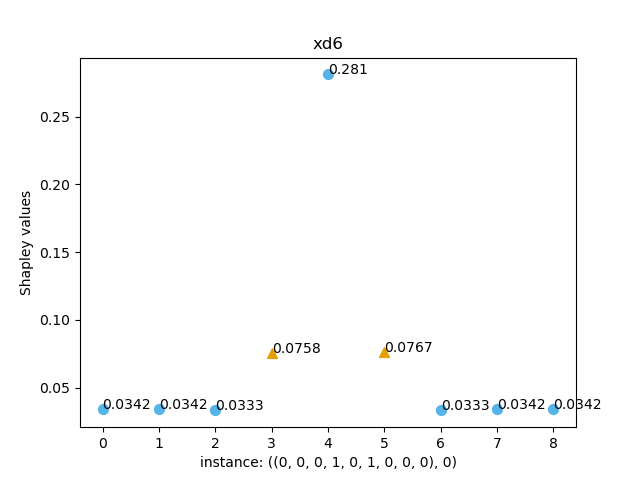}
    \caption{%
      Shapley values of a concrete instance. Shapley values of
      irrelevant features are shown as a yellow triangle, while
      Shapley value of relevant features are shown as a blue dot.} 
    \label{fig:sv_inst}
  \end{center}
\end{figure}

\begin{figure}[ht]
  \begin{center}
    \includegraphics[width=.475\textwidth]{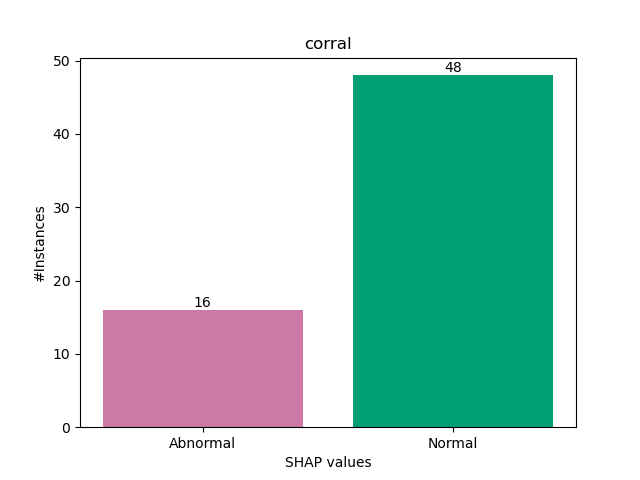}
    \quad
    \includegraphics[width=.475\textwidth]{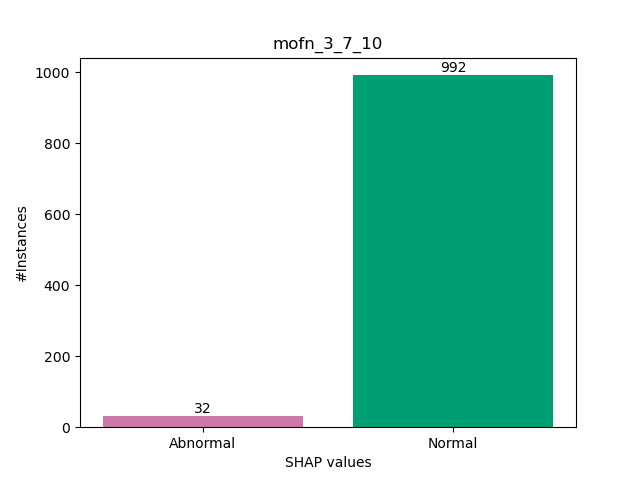} \\
    \includegraphics[width=.475\textwidth]{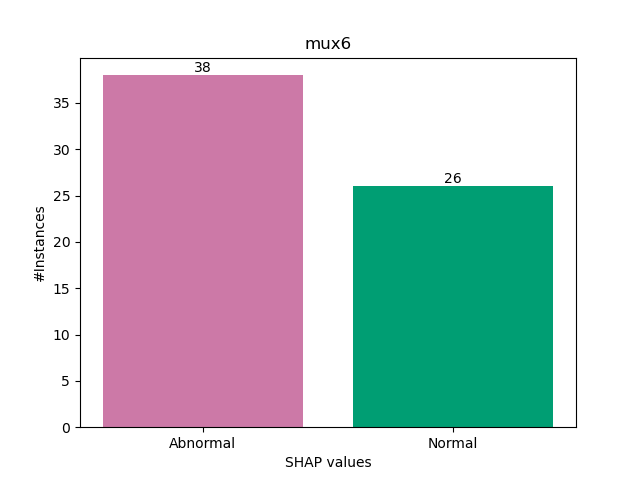}
    \quad
    \includegraphics[width=.475\textwidth]{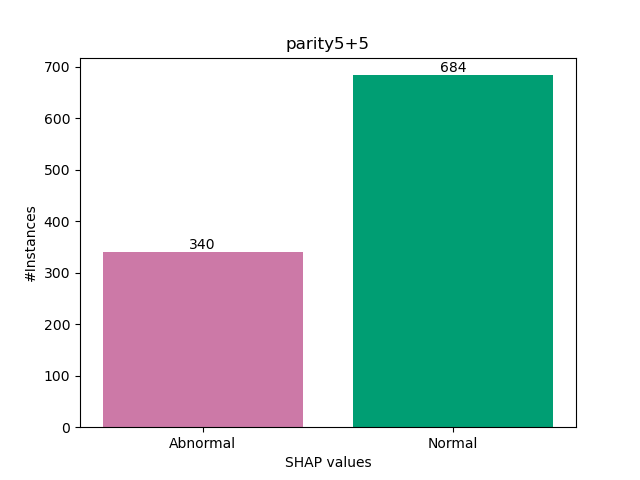} \\
    \includegraphics[width=.475\textwidth]{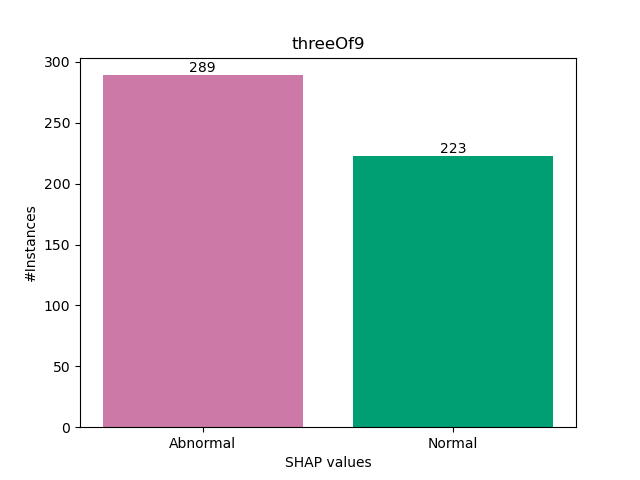}
    \quad
    \includegraphics[width=.475\textwidth]{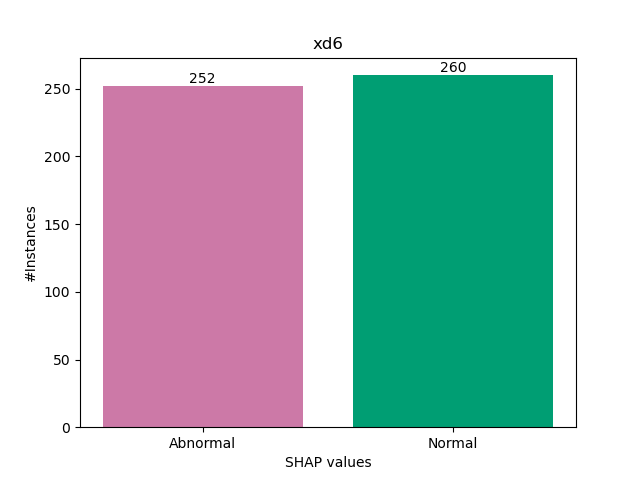}
    \caption{%
      Number of instances where the SHAP value of irrelevant features
      exceeds the minimum SHAP of relevant features (shown in pink),
      and number of instances where irrelevant features all have a
      score smaller than any relevant feature.
    }
    \label{fig:shap_irr_rel_diff}
  \end{center}
\end{figure}

\begin{figure}[ht]
  \begin{center}
    \includegraphics[width=.475\textwidth]{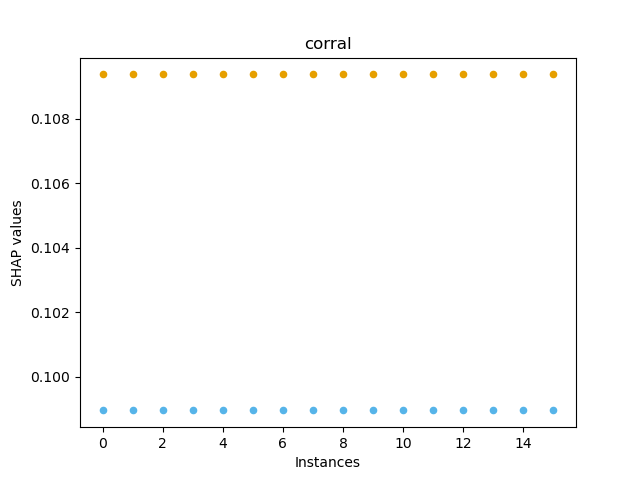}
    \quad
    \includegraphics[width=.475\textwidth]{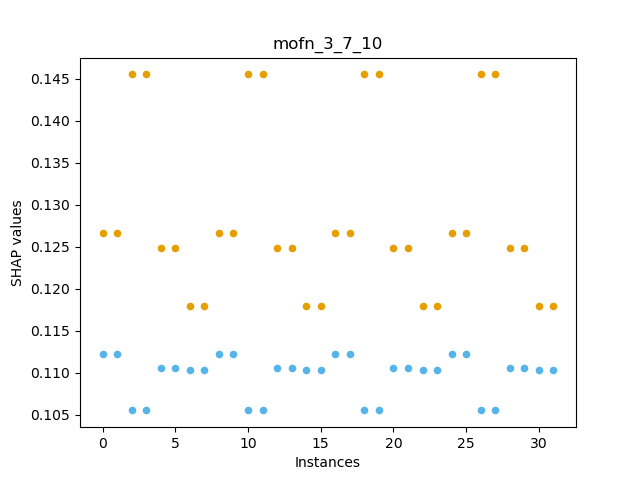} \\
    \includegraphics[width=.475\textwidth]{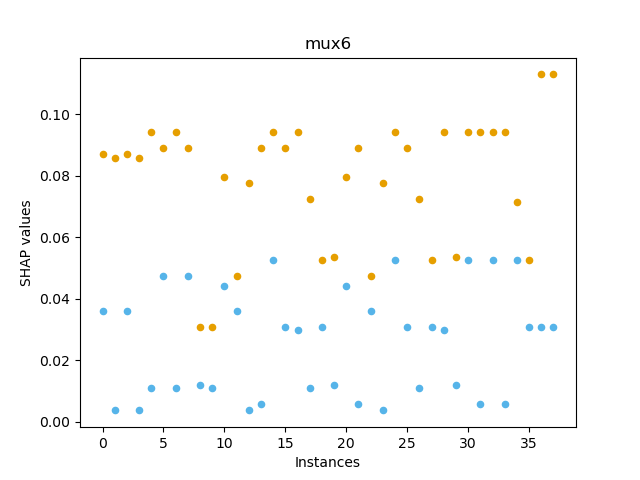}
    \quad
    \includegraphics[width=.475\textwidth]{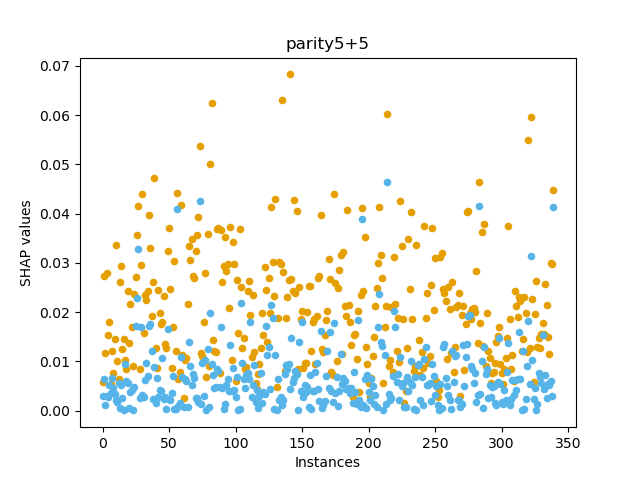} \\
    \includegraphics[width=.475\textwidth]{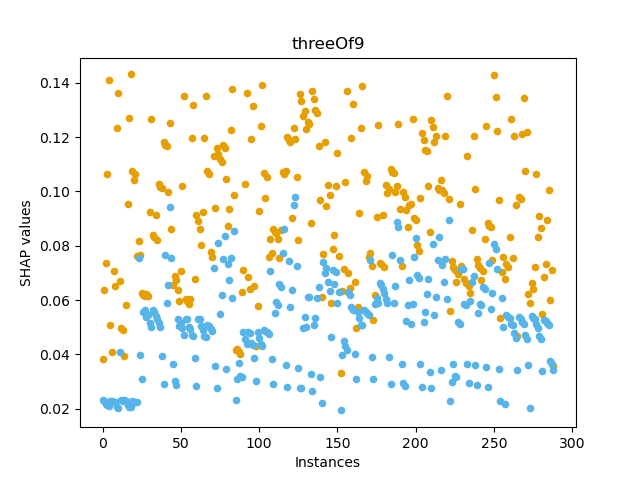}
    \quad
    \includegraphics[width=.475\textwidth]{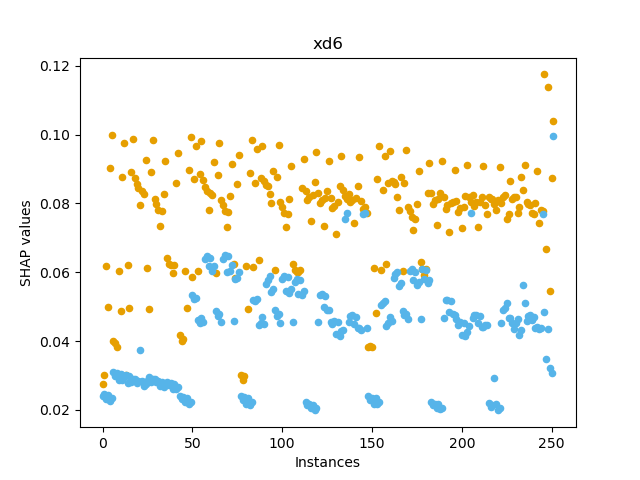}
    \caption{%
      Comparison of the maximum SHAP value of irrelevant features
      (dots in yellow) against the minimum SHAP value of relevant
      features (dots in blue). Only cases where there exist irrelevant
      features with higher scores than relevant features are shown.
    }
    \label{fig:shap_irr_rel}
  \end{center}
\end{figure}

\begin{figure}[ht]
  \begin{center}
    \includegraphics[width=.475\textwidth]{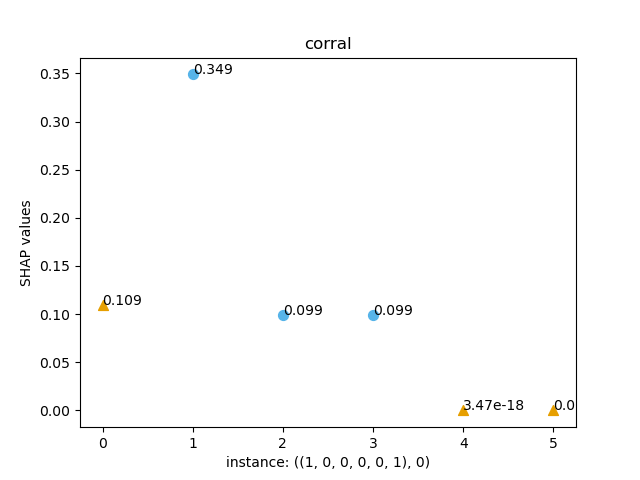}
    \quad
    \includegraphics[width=.475\textwidth]{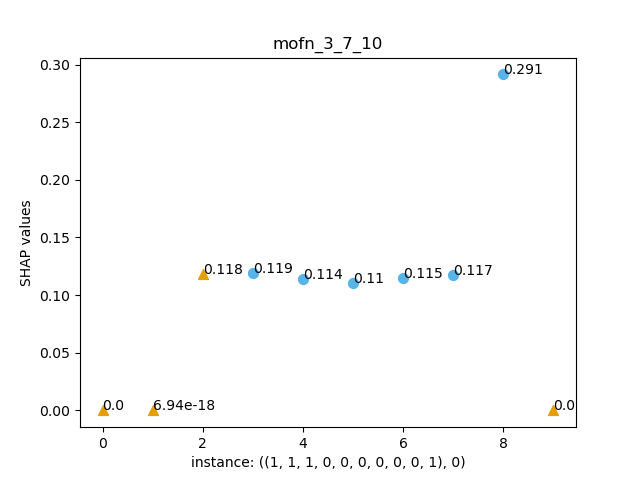} \\
    \includegraphics[width=.475\textwidth]{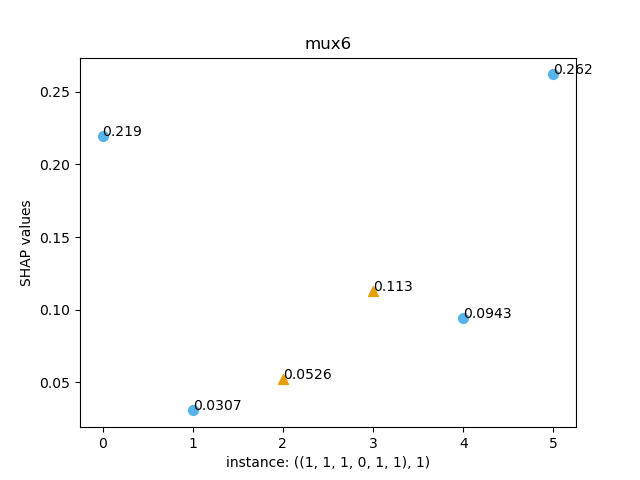}
    \quad
    \includegraphics[width=.475\textwidth]{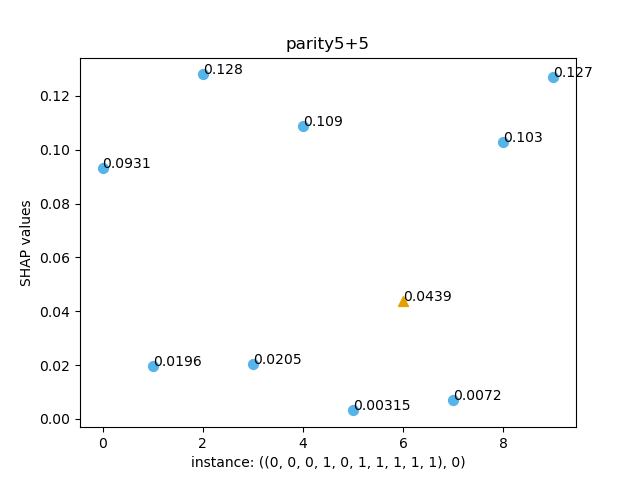} \\
    \includegraphics[width=.475\textwidth]{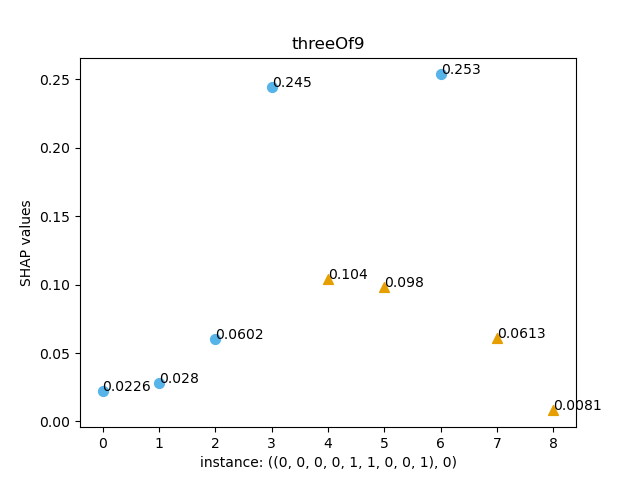}
    \quad
    \includegraphics[width=.475\textwidth]{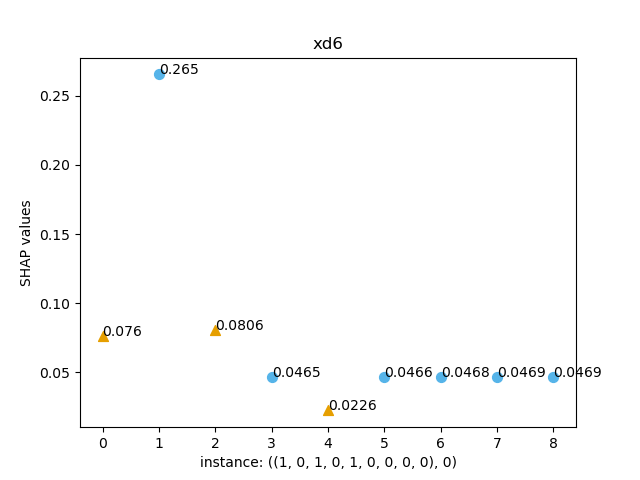}
    \caption{%
      SHAP values of a concrete instance. SHAP values of irrelevant
      features are shown as a yellow triangle, while SHAP values of
      relevant features are shown as a blue dot.}
    \label{fig:shap_inst}
  \end{center}
\end{figure}
\cref{fig:sv_irr_rel_diff,fig:sv_irr_rel} analyze each instance in
terms of whether the largest absolute Shapley value over all
irrelevant features exceeds the smallest absolute Shapley value over
all relevant features.
\cref{fig:shap_irr_rel_diff,fig:shap_irr_rel} repeats the exercise in
the case of values computed by SHAP.
Moreover, for each dataset, we picked a concrete instance to visualise
the difference of Shapley value between relevant features and
irrelevant features. This is shown in \cref{fig:sv_inst}.
One observation is that, with the exception of \texttt{parity5+5},
there exist irrelevant features can be declared as the $2^{\tn{nd}}$
or $3^{\tn{rd}}$ most important feature based on their Shapley value.
For \texttt{corral}, \texttt{mofn\_3\_7\_10}, \texttt{mux6},
\texttt{threeOf9} and \texttt{xd6}, it is simple to  be misled by the
computed exact Shapley values and conclude that an irrelevant feature
is important. This confirms that one cannot fully trust the
explanations based on Shapley values. 
For \texttt{parity5+5}, the Shapley value provides more trust since
most of the features are relevant and the Shapley value of the only
irrelevant feature one is smaller than several relevant features. 
Likewise, as depicted in \cref{fig:shap_inst}, we repeated such
analysis in the case of SHAP values. As can be observed, such problem
also exists for SHAP.

\end{document}